\documentclass{article} 
\usepackage{iclr2027_conference_arxiv,times}

\usepackage{amsmath,amsfonts,bm}

\def\eqref#1{equation~\ref{#1}}

\def\1{\bm{1}}

\DeclareMathAlphabet{\mathsfit}{\encodingdefault}{\sfdefault}{m}{sl}
\SetMathAlphabet{\mathsfit}{bold}{\encodingdefault}{\sfdefault}{bx}{n}

\usepackage[utf8]{inputenc} 
\usepackage[T1]{fontenc}    
\usepackage{hyperref}       
\usepackage{url}            
\usepackage{booktabs}       
\usepackage{amsfonts}       
\usepackage{nicefrac}       
\usepackage{microtype}      
\usepackage{xcolor}         
\usepackage{natbib}
\usepackage{amsmath}
\usepackage{algorithm}
\usepackage{algpseudocode}
\usepackage{hyperref}
\usepackage{url}
\PassOptionsToPackage{numbers}{natbib}
\usepackage{microtype}
\usepackage{graphicx}
\usepackage{subfigure}
\usepackage{booktabs} 
\usepackage{multirow}
\usepackage{pifont}
\usepackage{enumitem}
\usepackage{amssymb}
\usepackage{mathtools}
\usepackage{amsthm}
\usepackage{wrapfig}

\usepackage{booktabs}
\usepackage{adjustbox}
\usepackage{multicol}
\usepackage{multirow}
\usepackage{xspace}
\usepackage{algorithm}
\usepackage{algpseudocode}
\usepackage{enumerate}
\usepackage{enumitem}
\usepackage[most]{tcolorbox}
\tcbuselibrary{listings,skins}

\newtcblisting{promptbox}[1][System Prompt]{
  listing only, width=\textwidth,
  colback=gray!4, colframe=black!72, coltitle=white, fonttitle=\bfseries\small,
  title=#1, boxrule=0.6pt, arc=2pt, left=5pt, right=5pt, top=3pt, bottom=3pt,
  listing options={
    basicstyle=\ttfamily\footnotesize, breaklines=true, columns=fullflexible,
    keepspaces=true, extendedchars=true,
    literate={—}{{\textemdash}}1 {–}{{\textendash}}1 {“}{{``}}1 {”}{{''}}1 {’}{{'}}1,
  },
}

\newcommand{\cmark}{\textcolor{green!45!black}{\ding{51}}}
\newcommand{\xmark}{\textcolor{red!60!black}{\ding{55}}}
\newcommand{\winbadge}{\colorbox{green!50!black}{\textcolor{white}{\scriptsize\bfseries\,WIN\,}}}
\newcommand{\failbadge}{\colorbox{red!60!black}{\textcolor{white}{\scriptsize\bfseries\,FAILURE\,}}}
\newcommand{\thk}[1]{{\footnotesize\textcolor{violet!75!black}{\textbf{[think]}}\ \textit{#1}\par}}
\newcommand{\cmd}[1]{{\footnotesize\texttt{\textcolor{blue!35!black}{\$\ #1}}\par}}
\newcommand{\retr}[1]{{\footnotesize\textcolor{black!65}{\quad$\hookrightarrow$\ #1}\par}}
\newcommand{\ans}[1]{{\footnotesize\textbf{answer:}\ #1\par}}
\newtcolorbox[auto counter]{cmpbox}[1][]{%
  colback=black!2,colframe=black!55,boxrule=0.5pt,
  arc=2pt,left=4pt,right=4pt,top=3pt,bottom=3pt,fonttitle=\normalsize,
  before title={\textbf{Example~\thetcbcounter.}\ \ },#1}

\newtcolorbox[auto counter]{sftbox}[1][]{colback=black!2,colframe=green!45!black,
  boxrule=0.6pt,arc=2pt,left=4pt,right=4pt,top=3pt,bottom=3pt,fonttitle=\normalsize,
  before title={\textbf{SFT Example~\thetcbcounter.}\ \ },#1}

\newcommand{\ourmethod}{\texttt{GrepSeek}\xspace}

\title{\ourmethod{}: Training Search Agents for Direct Corpus Interaction}

\iclrfinalcopy

\author{
\textbf{Alireza Salemi\textsuperscript{1}}, \textbf{Chang Zeng\textsuperscript{1}}, \textbf{Atharva Nijasure\textsuperscript{1}}, \textbf{Jui-Hui Chung\textsuperscript{2}}, \textbf{Razieh Rahimi\textsuperscript{1}},\\
\textbf{Fernando Diaz\textsuperscript{3}}, \textbf{Hamed Zamani\textsuperscript{1}} \\
  \vspace{0.25cm}
  \textsuperscript{1}University of Massachusetts Amherst \,
  \textsuperscript{2}Princeton University \, \textsuperscript{3}Carnegie Mellon University \\
  \vspace{0.15cm}
  \texttt{\{asalemi,changzeng,anijasure,rahimi,zamani\}@cs.umass.edu} \\
  \texttt{juihui@princeton.edu} \quad \texttt{diazf@cmu.edu}
}

\begin{document}

\maketitle
\vspace{-0.4cm}
\begin{abstract}
\vspace{-0.4cm}
Large Language Model (LLM) search agents have shown strong promise on knowledge-intensive tasks through iterative reasoning and retrieval. Most existing systems rely on retrievers that return ranked documents from a pre-built index. We explore a complementary paradigm in which the agent treats the corpus as the search environment and finds evidence through executable shell commands. We introduce \ourmethod{}, an optimized direct corpus interaction (DCI) agent that learns to find, filter, and compose evidence over large text corpora. To stabilize reinforcement learning (RL) over large corpora, we train in two stages: first, we initialize the policy using verified, causally grounded search trajectories generated by an answer-aware Tutor and an answer-blind Planner; then, we refine the policy using Group Relative Policy Optimization (GRPO). To make DCI practical at scale, we introduce two semantics-preserving execution optimizations: Pruned Adaptive Command Execution, which reduces shell-based search latency by up to $77\times$ on a 14GB corpus with 21 million documents using a compact auxiliary structure, and Sharded-Parallel Corpus Search, which achieves up to $7.6\times$ speedup without additional preprocessing; both preserve equivalence with sequential execution. Across eight open-domain QA benchmarks, \ourmethod{} achieves the strongest overall performance, with a statistically significant relative improvement of $5.7\%$ over the best baseline. Our analysis shows how DCI-optimized agents conduct flexible and effective compositional search through direct corpus interaction.
\end{abstract}

\vspace{-0.8cm}
\section{Introduction}

LLM search agents \citep{search-o1,search-r1} have shown promise in addressing complex information needs that require reasoning, query decomposition, and/or information synthesis from multiple sources. These agents benefit from multiple interactions with a retriever to obtain information for performing their tasks. With unstructured or semi-structured text corpora, interactions are typically in the form of keyword or natural language queries. These paradigms rely on decades of research in developing retrieval models, from lexical \citep{TFIDF,bm25,QL} to semantic scoring based on dense \citep{LSI,karpukhin2020dense} or sparse representations \citep{SNRM,SPLADE}. They all operate on precomputed representations of documents to construct an index for the corpus.
Relevance scores between the query and documents are also mostly computed per document.

This paper explores a fresh perspective to this problem: \textit{information seeking on unstructured data can be done at any granularity}. In other words, instead of document-level representations, indexing, and relevance scoring, a piece of text of any size can be retrieved for each query. This enables us to perform more \textit{``surgical''} retrieval as opposed to being restricted with pre-determined text chunks and representations. 
To do this, \textit{we envision search agents that treat the corpus as an environment. Under this view, the agent can issue executable (surgical) search operations over the corpus, inspect results, refine constraints, and compose evidence across multiple steps}. This shifts retrieval from a black-box ranking procedure to an explicit sequence of controllable corpus operations. Such an interface is especially appealing for many knowledge-intensive reasoning tasks in which answering a question may require exact entity matching, lexical filtering, symbolic pattern search, or following bridge entities across documents. This perspective is closely related to recent progress in code agents, where executable search tools such as \texttt{grep} provide a simple yet effective interface for locating relevant context in code repositories \citep{wang2026grepragempiricalstudyoptimization}. Inspired by this form of tool-mediated search, we ask whether a similar interaction pattern can be extended beyond code repositories to open-domain QA over large textual corpora containing millions of unstructured documents.

Contemporary to our work, \citet{sen2026grepneedagentharnesses} and \citet{dci-agent} independently propose prompting methods to bypass precomputed retrieval indices and search a raw corpus through Unix-style shell commands. These works demonstrate that DCI can serve as an effective interface for exact matching, multi-step evidence discovery, and compositional QA. These methods are primarily built around prompting large proprietary models with strong code-generation capabilities to orchestrate search at inference time. For example, \citeauthor{dci-agent} rely on closed-weight agents such as Claude,
making the resulting system computationally expensive and operationally inefficient, often requiring substantial time, sometimes even one hour or more, to complete a single query \citep{dci-agent}. In contrast, we are interested in methods that are feasible in the real world, thus focusing on training compact models and efficient operation executions at large scale. In order to be consistent with these contemporary work, we also refer to this category of approaches as \textit{Direct Corpus Interaction (DCI)}.


\begin{wrapfigure}{r}{0.55\columnwidth}
    \centering
    \vspace{-0.6cm}
    \includegraphics[width=\linewidth]{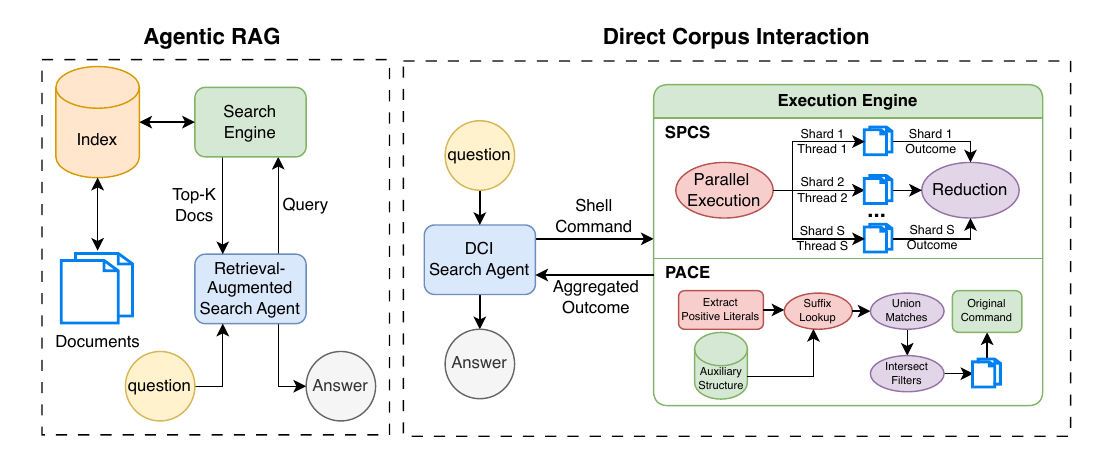}
    \vspace{-0.9cm}
    \caption{Comparing retrieval-based search and DCI. 
    }
    \label{fig:main-fig}
    \vspace{-0.3cm}
\end{wrapfigure}

To achieve effective and efficient DCI, we introduce \ourmethod{}, an optimized compact DCI agent that learns to search, filter, and compose evidence over large corpora with shell commands. This shifts DCI from inference-time prompting with proprietary LLMs (as in \citep{dci-agent}) to a learned capability of a smaller agent. Training DCI agents is challenging: naively applying RL often produces degenerate behavior, such as overly broad commands, excessive context retrieval, or unstable search behavior. To stabilize it, we first create a cold-start dataset that demonstrates successful DCI behavior. For each training question, an answer-aware Tutor model is given the ground-truth answer and constructs a backward chain of shell commands whose execution retrieves documents supporting the answer. This backward construction is useful for multi-hop questions, as it lets the Tutor model identify supporting evidence one step at a time while maintaining an explicit chain from the final answer back to the question. We then convert the backward chain into a forward, causally valid trajectory using an answer-blind Planner. The Planner generates reasoning traces and commands from the agent's observable history, simulating how the agent would solve the task at inference. The Tutor then aligns the steps with the verified backward chain. This produces trajectories that are causally grounded in the information observed so far, while allowing the Tutor model to guide the Planner towards the verified search path. Fine-tuning on the created cold-start dataset is followed by RL optimization using GRPO \citep{grpo}, allowing the DCI agent to improve its search behavior through direct interaction with the corpus.

For DCI to be practical, retrieval latency must remain manageable even when operating over millions of documents. However, executing shell pipelines sequentially over large multi-gigabyte collections introduces substantial I/O and processing bottlenecks, making naive execution slow for agents. We address this with two complementary semantics-preserving optimizations: Pruned Adaptive Command Execution (PACE), which narrows compatible searches to a conservative set of candidate documents before applying the original command, and a sharded-parallel execution engine that distributes compatible shell pipelines across corpus shards. Together, these mechanisms substantially reduce retrieval latency while preserving byte-exact equivalence with sequential execution.


\begin{wrapfigure}{r}{0.42\textwidth} 
    \centering
    \vspace{-0.6cm}
    \includegraphics[width=\linewidth]{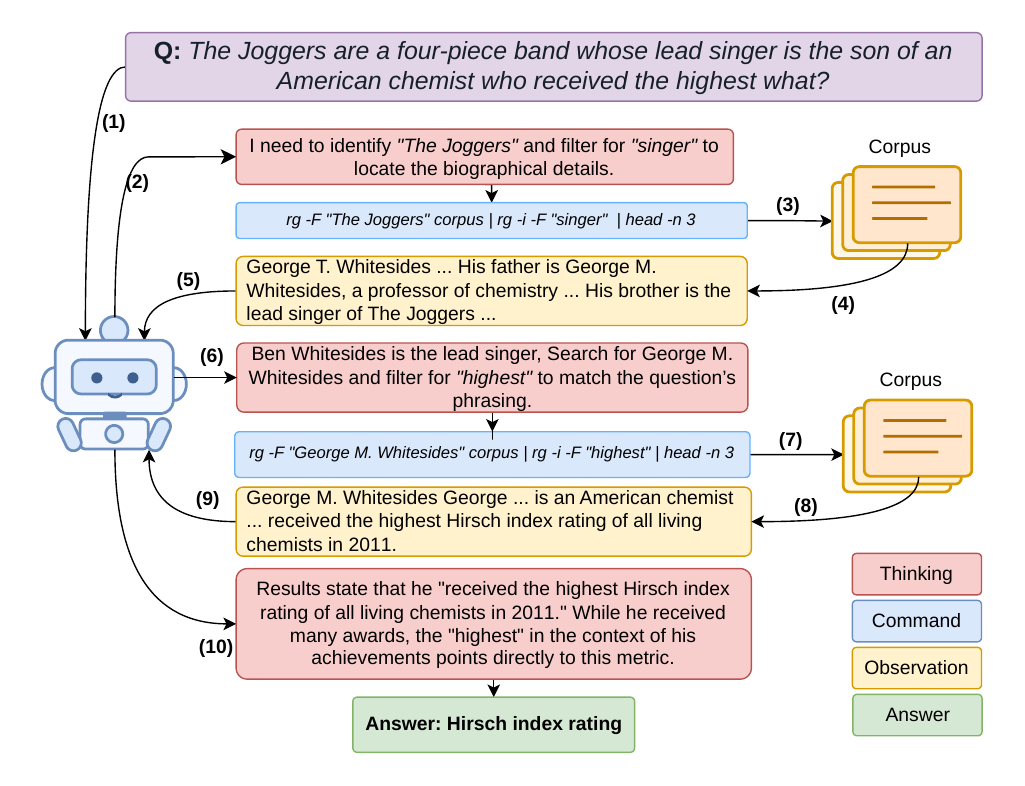}
    \vspace{-1cm}
    \caption{DCI workflow of \ourmethod{}.}
    \label{fig:dci-example}
    \vspace{-0.4cm}
\end{wrapfigure}
To evaluate \ourmethod{}, we conduct experiments across eight knowledge-intensive QA benchmarks: Natural Questions (NQ) \citep{nq}, TriviaQA \citep{triviaqa}, and PopQA \citep{popqa} for single-hop QA in addition to HotpotQA \citep{hotpotqa}, 2WikiMultiHopQA \citep{2wiki}, MuSiQue \citep{musique}, Bamboogle \citep{bamboogle}, and BrowseComp-Plus \citep{chen2025browsecompplusfairtransparentevaluation} for multi-hop QA that need iterative evidence aggregation and reasoning.
Experiments show that \ourmethod{} outperforms index-based RAG methods, untrained agentic frameworks, and even search agents optimized with RL to retrieve using dense and sparse retrievers. 
On average, \ourmethod{} achieves the best overall performance, with a $5.7\%$ statistically significant relative improvement over the strongest baseline, demonstrating DCI as a competitive and practical alternative to search agents that rely on index-based retrieval. 
Notably, it achieves the best performance on five out of eight benchmarks---NQ, HotpotQA, 2WikiMultiHopQA, MuSiQue, and BrowseComp-Plus.
The gains are pronounced on multi-hop tasks, where traditional retrieval systems frequently suffer from semantic conflation and entity ambiguity introduced by retrievers. 
We also show that our solutions are efficient for real-world applications. On a 14GB corpus containing 21 million documents, PACE reduces average retrieval latency from $5.39$ to $0.07$ seconds per query without sharding, yielding a $77\times$ speedup, at the cost of approximately $21$ minutes of one-time preprocessing to construct its 2.8GB auxiliary structure. In contrast, our sharded-parallel execution requires essentially no additional preprocessing and reduces latency to $0.71$ seconds with $32$ shards, yielding a $7.6\times$ speedup. Thus, PACE is preferable when one-time preprocessing is affordable, while sharded execution provides substantial acceleration when additional preprocessing is undesirable.
We will publicly release our codebase, the cold-start data, and model checkpoints.
\footnote{Available at: \url{https://github.com/alirezasalemi7/grepseek}}

\section{Optimizing Direct Corpus Interaction Search Agents}
\label{sec:method}

Concurrent and independent of our work, \citet{dci-agent} introduce DCI, where an agent bypasses precomputed retrieval indices and searches the corpus by Unix commands. This treats DCI as a prompting strategy, using strong code-generation LLMs such as Claude, making search computationally expensive and slow, taking up to an hour per query. Conversely, we study how to train smaller search agents to learn DCI capability, enabling effective corpus interaction through learned behavior.

\paragraph{DCI Search Agent:}

Figure~\ref{fig:main-fig}(Right) provides an overview of the DCI agent and Figure~\ref{fig:dci-example} shows a trajectory. DCI search agent $\pi_{\theta}$ operates under ReAct \citep{yao2023react}. Given a question $q$ and the system prompt in Figure~\ref{fig:sys-prompt} in Appendix~\ref{app:our-implementation}, the agent interacts directly with a corpus $\mathcal{C}$,\footnote{The corpus need not be stored as a single physical file. We expose it to the agent as one logical file for simplicity, while the execution engine maps commands to the underlying files and executes them accordingly.} where each line corresponds to a document. The interaction proceeds for at most $T$ steps, producing a trajectory $\tau = \{(t_i, a_i, o_i)\}_{i=1}^{T}$, where $t_i$ denotes the reasoning trace, $a_i$ the action, and $o_i$ the observation. At step $i$, conditioned on $q$ and the previous actions and observations $\tau_{<i}$, the policy $\pi_{\theta}$ generates a reasoning trace and an action: $(t_i, a_i) \sim \pi_{\theta}(\cdot \mid q, \tau_{<i})$. Reasoning traces are generated within \texttt{<think>} XML tags. Actions corresponding to tool invocations are emitted using the Hermes-style \texttt{<tool\_call>} format, and the tool outputs are returned to the model within \texttt{<tool\_response>} tags. When the agent decides to terminate, it produces the answer $\hat{y}_q$ within \texttt{<answer>} tags. The action $a_i$ is either a corpus interaction command\footnote{The action space consists of Unix tools, such as \texttt{rg}, \texttt{grep}, \texttt{find}, \texttt{sed}, \texttt{awk}, \texttt{head}, \texttt{tail}, \texttt{cat}, \texttt{ls}, \texttt{wc}, \texttt{sort}, \texttt{cut}, \texttt{uniq}, and \texttt{tr}. In practice, the agent primarily relies on \texttt{rg} and \texttt{head}.} (i.e., a shell command) or a termination that outputs the answer. Following each command, an execution engine runs the command over the corpus $\mathcal{C}$ and returns an observation $o_i$, which is appended to the trajectory and used for subsequent reasoning and action generation. The remainder of this section describes the training and efficient tool execution.

\subsection{Training DCI Search Agent}

We find that directly optimizing corpus interaction with RL is unstable: the agent generates ineffective commands and retrieves excessively large corpus segments, increasing context length and destabilizing training.\footnote{This frequently caused VRAM and host RAM OOM failures, even on systems with up to 1024\,GB of RAM.} We adopt a two-stage procedure. First, we construct a cold-start dataset to teach effective DCI behavior. We then supervise the model on this data before RL optimization.

\subsubsection{Cold-Start Data Generation}

Given a dataset $D=\{(q_i, y_i)\}_{i=1}^{|D|}$  of question-answer pairs requiring information from the corpus $\mathcal{C}$, our data pipeline (Figure~\ref{fig:data-overview} and Algorithm~\ref{alg:coldstart} in Appendix~\ref{app:our-implementation}) consists of two main phases followed by a filtering stage. The process uses an answer-aware Tutor ($\mathcal{M}_{T}$) to construct verified evidence chains, and an answer-blind Planner ($\mathcal{M}_{P}$) to synthesize realistic forward reasoning trajectories.

\begin{figure}
    \centering
    \includegraphics[width=0.85\linewidth]{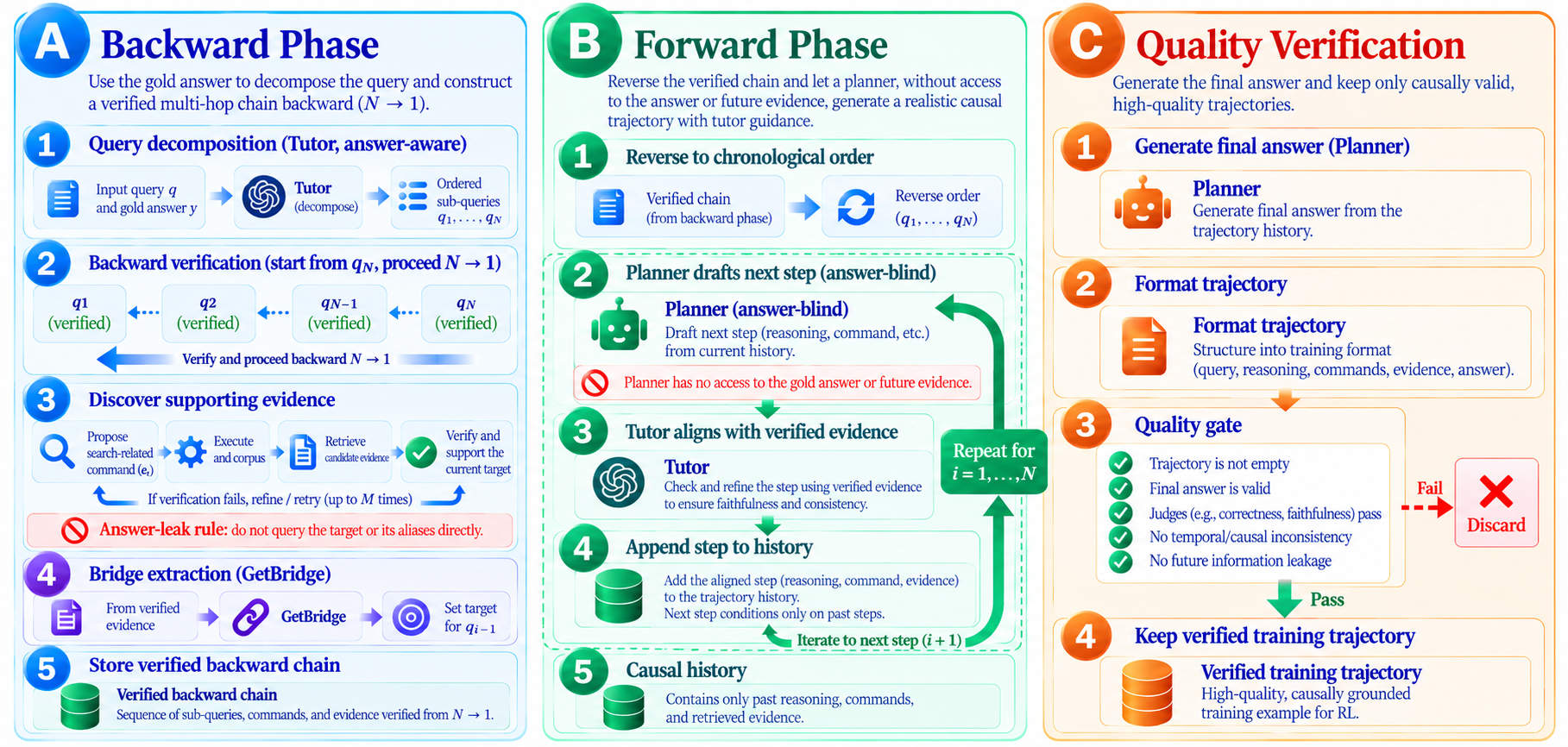}
    \vspace{-0.5cm}
    \caption{Overview of the cold-start data generation pipeline for DCI.}
    \label{fig:data-overview}
    \vspace{-0.6cm}
\end{figure}

\paragraph{Backward Phase:}

The Tutor decomposes the query $q$ and answer $y$ into a sequence of sub-queries (Algorithm~\ref{alg:coldstart}, line~\ref{line:decompose}; prompt in Figure~\ref{fig:prompt_decomposition} in Appendix~\ref{app:our-implementation}). To ensure the agent learns genuine information-seeking behavior rather than exploiting access to the answer, we construct the retrieval trajectory in reverse ($N \rightarrow 1$; lines~\ref{line:backward_loop}--\ref{line:bridge}). At each backward step, the Tutor proposes a shell command $c_i$ to retrieve a document $d_i$ that entails the current target answer $a$ (lines~\ref{line:discover} and~\ref{line:propose}). Crucially, we enforce an \textit{answer-leak rule} during command generation (prompt in Figure~\ref{fig:prompt_backward_system} in Appendix~\ref{app:our-implementation}): the proposed command must be target-masked, forbidding the use of the target $a$ or any of its aliases $F$ as retrieval terms. This is necessary as the backward process has access to future information through $y$; without masking, it can retrieve evidence by querying the answer, resulting in unrealistic retrieval behavior that doesn't reflect causal inference.
To improve retrieval robustness, the Tutor is allowed up to $M$ attempts (lines~\ref{line:refine_loop}--\ref{line:return_fail}). At each attempt, the Tutor proposes a command, executes it, and verifies whether retrieved documents support the answer using a verification step (line~\ref{line:exec_check}; prompt in Figure~\ref{fig:prompt_judge} in Appendix~\ref{app:our-implementation}). This increases the likelihood of obtaining valid evidence while filtering out brittle or spurious retrieval trajectories. Once at least one valid document is identified, a bridge extraction step determines the antecedent entity in $d_i$ that answers the preceding sub-query $q_{i-1}$ (lines~\ref{line:if_bridge}--\ref{line:bridge}; prompt in Figure~\ref{fig:prompt_bridge_extraction} in Appendix~\ref{app:our-implementation}). The extracted entity then becomes the answer for the next backward hop. After completing all steps, it produces a multi-hop chain connecting the original query to the final answer through a set of shell command steps.

\paragraph{Forward Phase:}

Upon finding a verified chain of documents and commands, the sequence is reversed  (Algorithm~\ref{alg:coldstart}, line~\ref{line:reverse_init}) to simulate the chronological information flow available to the agent in inference. Although the chain is created backward for verification purposes, a deployed agent only observes past interactions and retrieved evidence when making decisions. Reversing the trajectory ensures training trajectories faithfully match the inference behavior. At each forward step, the answer-blind Planner drafts a reasoning trace $\theta_d$ and action conditioned solely on the current causal history $\mathcal{H}$ (line~\ref{line:draft}; prompt in Figure~\ref{fig:prompt_planner_system} in Appendix~\ref{app:our-implementation}). Because the Planner does not have access to the verified evidence chain or future retrieval states, its proposed reasoning often lacks the precision necessary to justify the optimal shell command $c_i$. To bridge this gap, the Tutor model performs an alignment step that edits the Planner's reasoning trace to logically motivate $c_i$ while remaining grounded in the observable history (line~\ref{line:align}; prompt in Figure~\ref{fig:prompt_tutor_edit} in Appendix~\ref{app:our-implementation}). The resulting trajectory combines the realism of forward causal reasoning with the reliability of backward-verified evidence.

\paragraph{Filtering:}

To ensure the dataset provides a stable behavior for RL, it undergo a filtering (Algorithm~\ref{alg:coldstart}, lines~\ref{line:answer}--\ref{line:judge_gate}). First, the Planner generates a final answer $\hat{y}$ from the history $\mathcal{H}$ that achieves non-zero token-level overlap with the ground-truth $y$ ($\mathrm{F_1}(\hat{y}, y) > 0$). This ensures trajectories contain enough information for answer generation. Second, the trajectories are evaluated by the Tutor for causal/logical errors (line~\ref{line:judge_gate}; prompt in Figure~\ref{fig:prompt_coherence_judge} in Appendix~\ref{app:our-implementation}). The judge enforces temporal boundaries, discarding trajectories whose reasoning or commands reveal entities or facts not yet observable in the agent's current history. This is because the backward process has access to future information via the gold answer and verified evidence chain, and without verification, subtle forms of future leakage may persist even under explicit answer masking. Examples of data are in Appendix~\ref{app:sft-examples}.

\subsubsection{Optimization of DCI Search Agent}

\paragraph{SFT on Cold-Start Data:}

After creating the data, we perform supervised fine-tuning on them. Each example consists of the full interaction, including reasoning, tool invocations, tool responses, and the final answer. The objective is to initialize the agent with stable retrieval and reasoning behavior before RL. In particular, SFT teaches the agent to produce concise and causally grounded search commands and avoid pathological retrieval behavior such as excessively broad corpus scans.

\paragraph{RL with GRPO:}

We optimize the SFT-trained policy with GRPO \citep{grpo}.\footnote{Our method supports any standard RL algorithm; we use GRPO for its memory efficiency.} For each query $q$, $\pi_\theta$ samples $n=5$ trajectories $\tau^{(i)} \sim \pi_\theta(\cdot\mid q)$, each containing reasoning, tool interactions, and a final answer. Each $\tau^{(i)}$ receives an answer reward
$R_{\mathrm{ans}}(\tau^{(i)})$ based on token-level F$_1$ \citep{rajpurkar-etal-2016-squad} between the predicted answer $\hat{y}^{(i)}$ and the gold answer set $\mathcal{Y}$. To enforce the expected format, we define a binary indicator $\phi(\tau^{(i)}) \in \{0,1\}$ that checks whether the trajectory satisfies the required structural constraints. The final trajectory reward is therefore
$R(\tau^{(i)}) = \phi(\tau^{(i)})\, R_{\mathrm{ans}}(\tau^{(i)})$,
which assigns zero reward to structurally invalid trajectories. Details of the reward are provided in Appendix~\ref{app:reward}. GRPO computes group-relative advantages as $A^{(i)}=(R(\tau^{(i)})-\mu_R)/(\sigma_R+\epsilon)$, where $\mu_R$ and $\sigma_R$ denote the mean and standard deviation of rewards within the group. This encourages trajectories that outperform other samples for the same query. Initializing from the SFT policy improves optimization stability by providing structured retrieval behavior before RL.


\subsection{Efficient Corpus Interaction}
\label{sec:fast-retrieval}

Unlike RAG, which retrieves from a precomputed index, DCI searches the corpus directly by executing commands over millions of documents.\footnote{Our experiments use a 21M-passage Wikipedia corpus ($\sim$14GB).} The agent uses tools such as \texttt{rg}, \texttt{grep}, \texttt{sort}, \texttt{uniq}, \texttt{head}, and \texttt{tail}. Since a trajectory may require multiple corpus scans, execution efficiency is critical. We apply semantics-preserving optimizations that reduce command latency while receiving outputs identical to execution over the original corpus. We use two complementary strategies.


\paragraph{Pruned Adaptive Command Execution (PACE):} 

To reduce the cost of repeatedly scanning the full corpus, we introduce PACE, a semantics-preserving optimization inspired by inverted-index search \citep{InvertedIndexBook, InvertedFile}. PACE builds a lightweight auxiliary structure with two components: posting lists that map each vocabulary term to the documents containing it, and a suffix array \citep{suffix-array} that identifies vocabulary terms containing a given substring. To support substring matching in \texttt{rg}/\texttt{grep}, PACE first finds matching vocabulary terms through the suffix array, then combines their posting lists to identify candidate documents. This is important because shell search is not limited to exact token matches; for example, a query for \texttt{16} must also consider terms such as \texttt{2016} or \texttt{b16}. At inference time, PACE identifies filters that require documents to contain a specified string, such as \texttt{rg -F "London"}, which searches for the exact string \texttt{London}. Inverted filters, such as \texttt{rg -v "London"}, instead retain documents that do not contain the specified string. PACE excludes these filters from candidate construction because they do not identify a string that matching documents must contain. For each filter requiring a specified string, PACE splits that string into query terms. For each query term, two binary searches on the suffix array identify all vocabulary terms containing it as a substring. PACE then takes the \emph{union} of their posting lists to obtain candidate documents for that query term. These candidate sets are \emph{intersected} across query terms and filters, because a document must meet all required conditions to match the original command. The resulting candidate set contains every document that could match, but may also contain false positives. PACE reads the candidate documents from the original corpus and applies the original shell filters to determine the final results. The auxiliary structure determines only \emph{where} the command executes; the original filters determine which documents are returned. Commands that cannot be safely pruned fall back to full-corpus execution. Details of the auxiliary structure, candidate construction, and execution are provided in Appendix~\ref{app:eff-search-Pruned}.

\paragraph{Sharded-Parallel Corpus Search (SPCS):} 

To accelerate corpus interaction, we execute compatible shell pipelines in parallel over $S$ line-aligned corpus shards while preserving byte-exact equivalence with sequential execution (Algorithm~\ref{alg:sharded-search} in Appendix~\ref{app:eff-search-sharded}). The engine decomposes each pipeline, parallelizing it only when the first stage is a search operator (\texttt{rg} or \texttt{grep}) and all subsequent stages are shard-independent; otherwise, it falls back to sequential execution. Shard-local outputs are then reconstructed using operation-specific reductions: concatenation for stateless pipelines, local-plus-global truncation for \texttt{head -n}, summation for counts such as \texttt{wc -l}, and deterministic $k$-way merging for \texttt{sort}/\texttt{uniq} pipelines. This improves throughput without changing the agent's observations. Implementation details are provided in Appendix~\ref{app:eff-search-sharded}.

\section{Experiments}

\subsection{Experimental Setup}
\label{sec:exp-setup}

\paragraph{Datasets \& Evaluation:}

We evaluate on eight diverse QA datasets: three single-hop---NaturalQuestions (NQ) \citep{nq}, TriviaQA \citep{triviaqa}, and PopQA \citep{popqa}---and five multi-hop---HotpotQA \citep{hotpotqa}, 2WikiMultiHopQA (2Wiki) \citep{2wiki}, MuSiQue \citep{musique}, Bamboogle \citep{bamboogle}, and BrowseComp-Plus \citep{chen2025browsecompplusfairtransparentevaluation}. We obtain all datasets except BrowseComp-Plus from FlashRAG \citep{FlashRAG}. 
Following \citet{search-r1}, we use only NQ and HotpotQA for training, evaluating generalization on the remaining datasets. Dataset statistics are provided in Table~\ref{tab:dataset} in Appendix~\ref{app:dataset}. For the first seven datasets, we use the 2018 Wikipedia dump \citep{karpukhin2020dense} containing 21M passages.\footnote{Available at:\url{https://hf.co/datasets/PeterJinGo/wiki-18-corpus}, $\sim$14\,GB of text.} For BrowseComp-Plus, we use its accompanying corpus. We report token-level F$_1$ \citep{rajpurkar-etal-2016-squad} as the primary metric for the first seven datasets and the official script with GPT-4.1 \citep{openai2024gpt4technicalreport} judged accuracy for BrowseComp-Plus.

\paragraph{Training \& Inference Settings:}

We use Qwen3.5-9B \citep{qwen3.5} and Qwen3-4B \citep{yang2025qwen3technicalreport} as base models. For SFT, we construct a 10K-example dataset with a balanced mixture of NQ and HotpotQA. Qwen3.5-27B serves as both Tutor and Planner, with up to five refinement steps, and we train for one epoch. We optimize the policy with GRPO \citep{grpo} for 200 steps on NQ and HotpotQA using a group size of $n=5$. At inference, we use nucleus sampling \citep{Holtzman2020The} with temperature $0.6$. The agent is allowed up to $T=6$ turns and a 16{,}384-token context for FlashRAG datasets, and $T=16$ turns and a 131{,}072-token context for BrowseComp-Plus. Full SFT, GRPO, and inference configurations are provided in Tables~\ref{tab:hyperparams-sft}, \ref{tab:hyperparams-grpo}, and \ref{tab:hyperparams-inference} in Appendix~\ref{app:experimental-setting}.

\paragraph{Baselines:}

Following \citet{search-r1}, we compare against a range of baselines, including (1) a direct LLM inference without external search, and (2) retrieval-augmented methods: RAG \citep{rag}, IRCoT \citep{ircot}, Search-O1 \citep{search-o1}, rejection sampling with a search engine and Search-R1 \citep{search-r1} (GRPO-optimized).\footnote{While recent agentic frameworks improve reasoning \citep{jin2026beneficialreasoningbehaviorsagentic, sun-etal-2025-simpledeepsearcher} or multi-agent orchestration \citep{chen2026monolithicarchitecturesmultiagentsearch}, our goal is to isolate the effect of retrieval. We therefore compare methods that differ in how evidence is retrieved---traditional sparse and dense retrievers versus DCI. These orchestration improvements are largely orthogonal and could be incorporated into either paradigm.} All baselines use the same backbone LLMs and are trained (when applicable) and evaluated under the same settings as our method for a fair comparison (Appendix~\ref{app:experimental-setting}). We use three retrievers to retrieve top-3 documents from the corpus: BM25 \citep{bm25} as a sparse lexical baseline, E5 \citep{e5} as a dense embedding (110M parameters), and a large-scale Qwen3 embedding model \citep{qwen3embedding} (4B parameters). This allows us to systematically assess performance across both traditional and strong neural retrievers. Dense retrievers are implemented using FAISS \citep{douze2025faisslibrary} with HNSW index \citep{hnsw} ($M=32$, $\text{efConstruction}=128$, $\text{efSearch}=128$).

\subsection{Main Findings}

\begin{table*}[t]
\centering
\caption{Performance across datasets (Qwen3.5 backbone). We report $F_1$ for the FlashRAG datasets and GPT-4.1-judged accuracy for BrowseComp-Plus. EM is reported in Table~\ref{tab:em-result} in Appendix~\ref{app:more-results}. Superscript $^*$ denotes datasets used during training, while $^\uparrow$ and $^\downarrow$ denote statistically significant improvement and degradation, compared to the best baseline ($p<0.05$).}
\label{tab:f1-result}

\resizebox{\textwidth}{!}{
\begin{tabular}{llccccccccc}
\toprule
\textbf{Method} &
\textbf{Retriever} &
\multicolumn{3}{c}{\textbf{Single-hop}} &
\multicolumn{5}{c}{\textbf{Multi-hop}} &
\textbf{Average} \\
\cmidrule(lr){3-5}
\cmidrule(lr){6-10}
& &
\textbf{NQ$^*$} &
\textbf{TriviaQA} &
\textbf{PopQA} &
\textbf{HotpotQA$^*$} &
\textbf{2Wiki} &
\textbf{MuSiQue} &
\textbf{Bamboogle} &
\textbf{\shortstack{BrowseComp+}} &
\textbf{(macro)} \\
\midrule

Direct
& ---
& 0.2733
& 0.5565
& 0.2364
& 0.2837
& 0.3353
& 0.1151
& 0.1648
& 0.0036
& 0.2461 \\

\midrule

\multirow{3}{*}{RAG}
& BM25
& 0.3329
& 0.6660
& 0.3239
& 0.4434
& 0.3469
& 0.1305
& 0.2841
& 0.0916
& 0.3274 \\
& E5-110M
& 0.5068
& 0.7072
& 0.4468
& 0.4206
& 0.3227
& 0.1494
& 0.3383
& 0.0398
& 0.3664 \\
& Qwen3-4B
& 0.5002
& 0.7212
& 0.5046
& 0.4548
& 0.3498
& 0.1609
& 0.3484
& 0.0964
& 0.3920 \\

\midrule

\multirow{3}{*}{IRCoT}
& BM25
& 0.2607
& 0.5518
& 0.2757
& 0.3098
& 0.1356
& 0.0725
& 0.1494
& 0.0952
& 0.2313 \\
& E5-110M
& 0.4473
& 0.6238
& 0.3970
& 0.2983
& 0.1360
& 0.0921
& 0.2549
& 0.0277
& 0.2846 \\
& Qwen3-4B
& 0.4379
& 0.6509
& 0.4548
& 0.3464
& 0.1641
& 0.1210
& 0.2770
& 0.0867
& 0.3173 \\

\midrule

\multirow{3}{*}{Search-O1}
& BM25
& 0.4261
& 0.7283
& 0.4003
& 0.4893
& 0.3735
& 0.2271
& 0.5011
& 0.1867
& 0.4165 \\
& E5-110M
& 0.4771
& 0.7316
& 0.4322
& 0.4706
& 0.3558
& 0.2262
& 0.6158
& 0.0614
& 0.4213 \\
& Qwen3-4B
& 0.4622
& 0.7290
& 0.4731
& 0.4828
& 0.4009
& 0.2517
& 0.6103
& 0.0867
& 0.4371 \\

\midrule

\multirow{3}{*}{\shortstack[l]{Rejection\\Sampling}}
& BM25
& 0.3825
& 0.7228
& 0.3827
& 0.5530
& 0.3972
& 0.2396
& 0.5819
& 0.2651
& 0.4406 \\
& E5-110M
& 0.4238
& 0.7256
& 0.4298
& 0.5353
& 0.3678
& 0.2438
& 0.6441
& 0.1410
& 0.4389 \\
& Qwen3-4B
& 0.4294
& 0.7258
& 0.4630
& 0.5442
& 0.4255
& 0.2697
& 0.6569
& 0.1759
& 0.4613 \\

\midrule

\multirow{3}{*}{Search-R1}
& BM25
& 0.4592
& 0.7733
& 0.4194
& 0.5659
& 0.4028
& 0.2510
& 0.6013
& 0.2663
& 0.4674 \\
& E5-110M
& 0.5069
& \textbf{0.7734}
& 0.4747
& 0.5424
& 0.3727
& 0.2720
& 0.6763
& 0.1446
& 0.4704 \\
& Qwen3-4B
& 0.5067
& 0.7693
& \textbf{0.5101}
& 0.5591
& 0.4299
& 0.2878
& \textbf{0.6989}
& 0.1940
& 0.4945 \\

\midrule

\ourmethod
& ---
& \textbf{0.5223$^\uparrow$}
& 0.7673
& 0.4861$^\downarrow$
& \textbf{0.6231$^\uparrow$}
& \textbf{0.5178$^\uparrow$}
& \textbf{0.3006}
& 0.6212
& \textbf{0.3422$^\uparrow$}
& \textbf{0.5226$^\uparrow$} \\

\bottomrule
\end{tabular}
}

\vspace{-0.4cm}
\end{table*}

\paragraph{Comparison of Performance with Baselines:}

We compare \ourmethod{} against retrieval-augmented agentic search baselines in Table~\ref{tab:f1-result}. \ourmethod{} outperforms non-agentic approaches (Direct and RAG), untrained agents (IRCoT and Search-O1), and trained baselines (Rejection Sampling), with Search-R1 as the strongest competitor. Overall, \ourmethod{} achieves the best performance on 5 of 8 benchmarks, with statistically significant gains on 4 ($^\uparrow$), particularly on multi-hop tasks. These gains highlight the effectiveness of DCI for iterative evidence aggregation and entity-level precision, such as distinguishing subsidiaries from parent companies (Example~\ref{ex:hotpot-rockstar}) and avoiding name-collision errors (Example~\ref{ex:hotpot-faithlutheran}) in Appendix~\ref{app:case-studies}. Performance is slightly lower on TriviaQA and Bamboogle without statistical significance, while PopQA is the only dataset with a significant drop ($^\downarrow$). These trade-offs reflect the lexical nature of \ourmethod{}. DCI is highly effective when queries contain strong textual anchors, including rare formulas (Example~\ref{ex:hotpot-formula}), distinctive phrasing (Example~\ref{ex:hotpot-joggers}), and exact entity names (Example~\ref{ex:popqa-arndt}) in Appendix~\ref{app:case-studies}. However, limited lexical overlap and form variation remain challenging: on PopQA, diacritics can prevent exact matches (Example~\ref{ex:popqa-vaillant}), while overloaded keywords can cause \texttt{rg} to surface earlier but less authoritative matches because it lacks relevance ranking (Example~\ref{ex:bamboogle-citibank}). Dense retrievers are better suited to such cases by mapping lexical variants into a shared embedding space. Despite these limitations, \ourmethod{} achieves the strongest overall score ($0.5226$), significantly outperforming the best dense retrieval baseline ($p<0.05$), showing that DCI provides a precise and effective alternative for open-domain QA and multi-hop reasoning.

\begin{figure}[!t]
    \centering
    \includegraphics[width=\textwidth]{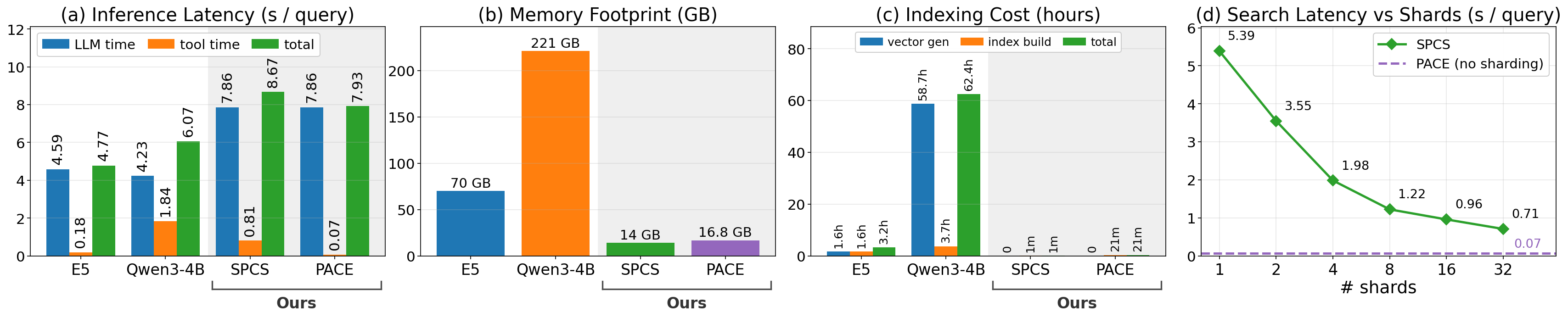}

    \vspace{-0.5cm}
    \caption{Efficiency and cost analysis of \ourmethod{} compared to dense retrieval baselines (E5 and Qwen3-4B). (a) Inference latency per query, broken down into LLM generation and tool execution time. (b) Memory footprint (RAM) required for the retrieval index. (c) Offline preprocessing time, measured in A100 GPU-hours for the dense retrievers and CPU wall-clock hours for SPCS and PACE. (d) Search tool latency of \ourmethod{} scaling with the number of shards.}
    \label{fig:efficiency}
    \vspace{-0.4cm}
\end{figure}

\paragraph{Comparison of Latency with Best-Performing Baselines:}

Based on Table~\ref{tab:f1-result}, Search-R1 with E5 and Qwen3 embeddings are the strongest dense retrieval baselines. We compare their efficiency with our two DCI execution strategies---SPCS and PACE---in Figure~\ref{fig:efficiency}. All inference experiments use 32 CPU cores over 50 examples from each FlashRAG dataset (350 total), while dense-retriever indexing uses a single NVIDIA A100 GPU. As shown in Figure~\ref{fig:efficiency}(a), SPCS has an end-to-end latency of $8.67$\,s per query, including $0.81$\,s of tool execution, while PACE reduces tool execution to only $0.07$\,s and total latency to $7.93$\,s. In comparison, E5 and Qwen3-4B require $4.77$\,s and $6.07$\,s, respectively; the remaining DCI overhead is primarily due to longer LLM reasoning trajectories. Figure~\ref{fig:efficiency}(b--c) highlights the memory and preprocessing trade-offs. SPCS requires only the $14$\,GB raw corpus, while PACE requires $16.8$\,GB, including its compact auxiliary structure. In contrast, E5 and Qwen3-4B require $70$\,GB and $221$\,GB, respectively, to store their retrieval structures. SPCS requires approximately $1$ minute of CPU wall-clock time for corpus setup, while constructing the PACE auxiliary structure takes approximately $21$ minutes of CPU wall-clock time. Both avoid expensive embedding generation, which requires $3.2$ and $62.4$ A100 GPU-hours for E5 and Qwen3-4B, respectively. Finally, Figure~\ref{fig:efficiency}(d) compares search latency across execution strategies. SPCS scales from $5.39$\,s with one shard to $0.71$\,s with 32 shards, with diminishing gains at higher shard counts due to system-level bottlenecks such as memory bandwidth and merging overhead. PACE, even without sharding, reduces average search latency to $0.07$\,s by restricting execution to candidate documents rather than scanning the full corpus. These results show that the two optimizations are complementary: SPCS accelerates full-corpus execution through parallelism, whereas PACE avoids most of the scan altogether through candidate pruning.

\begin{figure*}[t]
    \centering
    \includegraphics[width=0.72\textwidth]{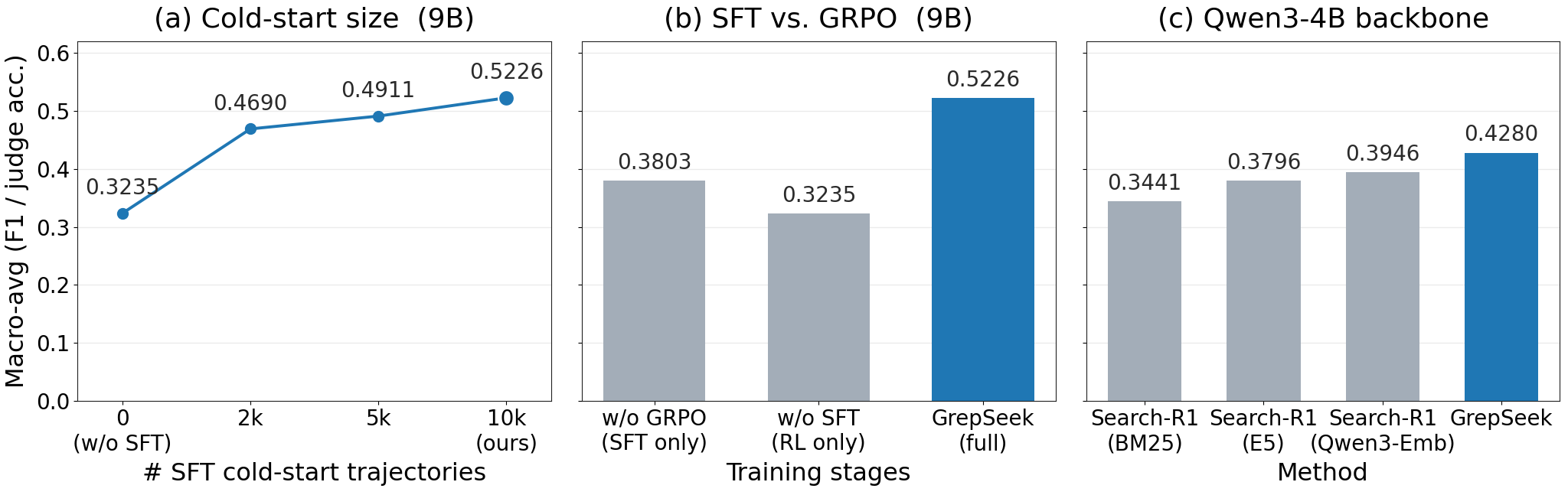}
    \vspace{-0.4cm}
    \caption{Ablation of \ourmethod{} using the average performance on all datasets.
    (a) Effect of cold-start SFT data size using the 9B backbone. 
    (b) Contribution of the SFT and GRPO training stages for the 9B model. 
    (c) Comparison of \ourmethod{} with Search-R1 variants using the Qwen3-4B backbone. Per-dataset results for these analyses are reported in Appendix~\ref{app:more-results}.}
    \vspace{-0.4cm}
    \label{fig:ablation_scaling}
\end{figure*}

\paragraph{Ablations of \ourmethod:}

We study the contribution of each training stage using macro-average performance in Figure~\ref{fig:ablation_scaling}(b), with full per-dataset $F_1$ and EM results reported in Tables~\ref{tab:ablation_f1} and~\ref{tab:ablation_em} in Appendix~\ref{app:more-results}.\footnote{As discussed earlier, directly optimizing the base model without SFT initialization was highly unstable. For the \textit{w/o SFT} setting, we therefore report results from the final checkpoint before training collapse.} The full \ourmethod{} reaches a macro-average score of $0.5226$, compared with $0.3803$ without GRPO and $0.3235$ without SFT, showing that both cold-start supervision and subsequent RL optimization are important. The larger degradation without SFT further highlights the importance of structured trajectory initialization before reinforcement learning. We further vary the cold-start set from $0$ to $2$k, $5$k, and $10$k trajectories while keeping the subsequent GRPO configuration fixed (Figure~\ref{fig:ablation_scaling}(a)). Macro-average performance increases from $0.3235$ without SFT to $0.4690$ with only $2$k trajectories, then improves more gradually to $0.4911$ and $0.5226$ with $5$k and $10$k trajectories, respectively. Finally, to test whether these gains generalize across model scales, we repeat the comparison using the smaller Qwen3-4B backbone (Figure~\ref{fig:ablation_scaling}(c)). \ourmethod{} again achieves the highest macro-average score ($0.4280$), outperforming Search-R1 with BM25 ($0.3441$), E5 ($0.3796$), and Qwen3 embeddings ($0.3946$). Per-dataset 4B results are provided in Table~\ref{tab:f1-result-4b} in Appendix~\ref{app:more-results}, indicating that the benefits of DCI persist across different backbone sizes.

\paragraph{Training Dynamics:}

\begin{figure}[!t]
    \centering
    \includegraphics[width=0.72\textwidth]{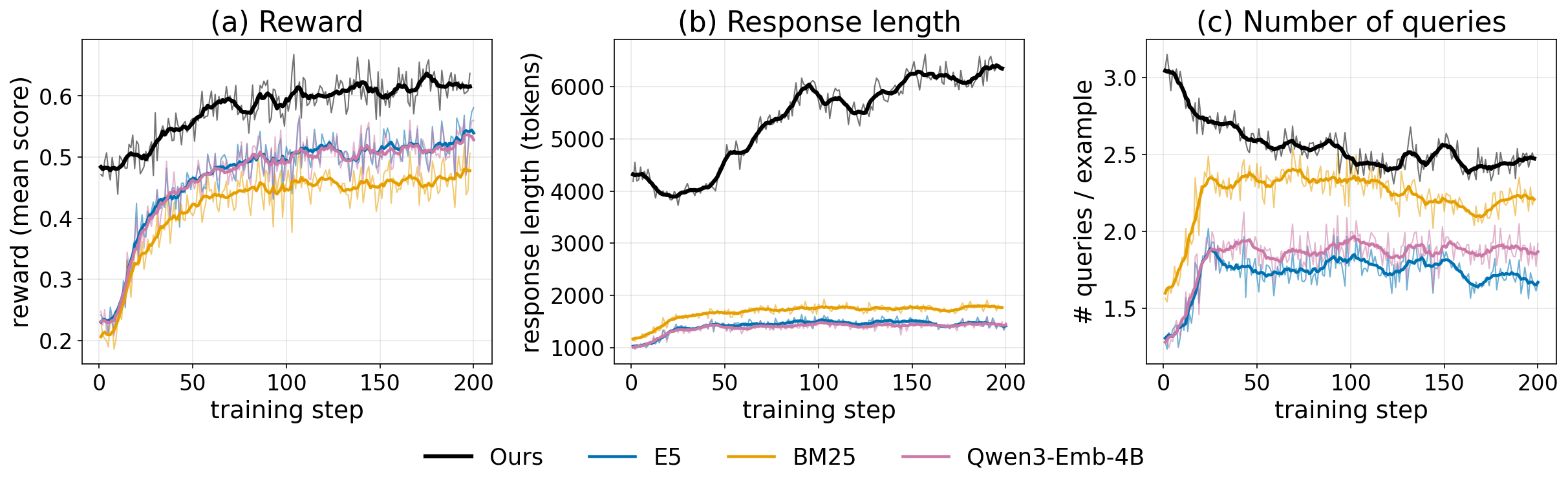}
    \vspace{-0.6cm}
    \caption{Training dynamics (Qwen3.5) over 200 steps comparing \ourmethod{} with retrieval baselines (E5, BM25, and Qwen3-Emb-4B). (a) Mean reward score during training. (b) Average response length measured in tokens. (c) Average number of search queries generated per example.}
    \vspace{-0.4cm}\label{fig:training_dynamics}
\end{figure}

Figure~\ref{fig:training_dynamics} shows the training dynamics over 200 GRPO steps. \ourmethod{} consistently achieves higher rewards than Search-R1 variants using E5, BM25, and Qwen3-Emb-4B (Figure~\ref{fig:training_dynamics}(a)), but generates longer sequences due to extended reasoning and raw corpus context, reducing throughput (Figure~\ref{fig:training_dynamics}(b)). Its retrieval behavior also evolves differently: while baselines increase their number of queries before stabilizing, \ourmethod{} gradually reduces the number of executed commands (Figure~\ref{fig:training_dynamics}(c)): the agent shifts from multiple independent searches toward more expressive multi-stage shell pipelines, extracting more information per command invocation.

\subsection{Analysis}

\paragraph{Retrieval Behavior:}

Because DCI interacts with the corpus with commands, its retrieval behavior is directly interpretable. Across the seven FlashRAG benchmarks, \ourmethod{} consistently uses \texttt{|\,head} to limit output, relies on fixed-string matching (\texttt{-F}) in nearly all searches, and uses cascaded filtering in 70\% of commands. It adapts search effort to task difficulty, issuing more commands on multi-hop datasets (2.8--3.8) than on single-hop datasets (2.3--2.7). Under the longer-horizon BrowseComp-Plus setting, which allows up to 16 turns, this increases to 6.9 commands per query. To separate behaviors learned during SFT from those refined by RL, we track trajectory statistics across training (Table~\ref{tab:cmd-evo} in Appendix~\ref{app:more-results}). Pipeline structure---pipe depth, fixed-string matching, truncation, and cascaded filtering---remains largely stable, suggesting these retrieval primitives are established during SFT. RL instead shapes higher-level behavior: the agent reduces commands per trajectory ($3.06 \rightarrow 2.56$), retrieves more context per query (\texttt{head -n} from roughly 5 to 10 lines), and allocates more tokens to reasoning ($4{,}251 \rightarrow 6{,}409$).

\paragraph{Case Studies:}

To qualitatively analyze \ourmethod{}, we compare its trajectories with the strongest baseline, Search-R1 with Qwen3-Emb-4B (Appendix~\ref{app:case-studies}). The traces highlight the lexical precision of DCI: \ourmethod{} can isolate rare symbolic patterns (Example~\ref{ex:hotpot-formula}), exact entity names (Example~\ref{ex:popqa-arndt}), compose keywords across multiple hops (Example~\ref{ex:hotpot-joggers}), and resolve entity collisions (Examples~\ref{ex:hotpot-rockstar} and \ref{ex:hotpot-faithlutheran}). However, lexical search has clear limitations. Since Unix tools return matches in file order rather than relevance, important evidence may be preceded by less informative passages (Example~\ref{ex:bamboogle-citibank}), while spelling variations can prevent a match (Example~\ref{ex:popqa-vaillant}). Dense retrieval can be more robust in these cases by generalizing across lexical variations in embedding space.

\section{Related Work}

A more detailed discussion is provided in Appendix~\ref{app:detailed-related-work}.

\paragraph{Retrieval-Augmented Agentic Search:}
RAG augments LLMs with external knowledge by retrieving relevant evidence from a corpus, typically through sparse or dense index-based retrieval \citep{rag,REML,bm25,e5,qwen3embedding}. For complex information needs, recent work extends this paradigm to multi-step search, where reasoning and retrieval are interleaved across multiple turns \citep{ircot,search-r1,search-o1,starqa}. Existing agentic-search methods primarily improve how an LLM plans, formulates queries, or interacts with a conventional retriever, with some methods additionally optimizing the retrieval component itself \citep{search-r1,search-o1,zeng2026cosearch}. In contrast, our work trains a compact open-weight model to interact directly with the corpus through deterministic shell-based search operations. We evaluate this capability on single-hop and multi-hop question answering, following established evaluation settings for retrieval-augmented search agents. 

\paragraph{Direct Interaction with Corpus:}
DCI bypasses conventional retrieval and allows an agent to explicitly search, filter, and compose evidence over the corpus. Earlier work studied direct textual search in code repositories and regular-expression matching over text collections \citep{10.1145/3565971,wang2026grepragempiricalstudyoptimization,188994,NAVARRO2003423}. More recent work has begun applying DCI-style interfaces to open-domain QA \citep{sen2026grepneedagentharnesses,dci-agent,subramanian2025keywordsearchneedachieving}. While these demonstrate that LLMs can use direct corpus operations, they focus on exposing capable proprietary models to a direct-search interface or designing the surrounding agent harness. Our work instead focuses on \emph{learning} this capability: we train a compact open-weight model to develop multi-step direct-search behavior, including how to formulate, sequence, and refine corpus operations during reasoning. Thus, rather than treating DCI primarily as an inference-time interface for an already capable agent, we study how direct corpus interaction itself can be acquired through training.

\section{Conclusion}

We introduced \ourmethod{}, a DCI agent that operates directly over raw corpora with Unix commands. Cold-start SFT and GRPO yield effective, interpretable, and lexically precise retrieval programs that lead overall across our QA benchmarks, particularly on multi-hop tasks. To make DCI efficient at scale, PACE narrows search to conservative candidate sets, while sharded parallelism accelerates full-corpus scans. Both preserve sequential semantics while offering complementary preprocessing--latency trade-offs. Our analysis also highlights the limitations of lexical retrieval, particularly sensitivity to surface-form variation and the lack of semantic relevance ranking. Future work will explore hybrid DCI and index-based retrieval, richer matching primitives such as fuzzy search, and more efficient reasoning and context management. We also plan to extend evaluation to long-form QA, ad hoc retrieval, and previously unseen corpora.

\section*{Acknowledgments}

This work was supported in part by the Center for Intelligent Information Retrieval, in part by NSF grant number 2402873, in part by the Office of Naval Research contract number N000142412612, and with support from Google.org. Any opinions, findings and conclusions or recommendations expressed in this material are those of the authors and do not necessarily reflect those of the sponsor.

\newpage
\subsection*{AI Use Statement}

Generative AI tools were used throughout the development of this work for several purposes. 
They were used to assist with writing and editing, including drafting and revising portions of the manuscript, improving clarity and presentation, and preparing and refining figures and visualizations. 
They were also used during software development to assist with code generation, debugging, implementation, and the development of experimental and analysis utilities. 
In addition, generative AI tools were used for research ideation and execution, including brainstorming methodological alternatives, discussing implementation choices, and assisting with the retrieval and discovery of relevant prior work. Generative models also form part of the research methodology itself. In particular, as described in the paper, we use language models to generate synthetic cold-start search trajectories through the Tutor--Planner pipeline. These trajectories are subsequently verified and filtered before being used for supervised fine-tuning. Generative AI was therefore used for synthetic data generation in addition to its use as a general research and writing assistant.

All AI-assisted writing was reviewed and edited by the authors. AI-suggested research ideas, technical claims, and related work were independently checked by the authors against the relevant literature and primary sources. AI-assisted code was reviewed, executed, and tested as part of the experimental pipeline, and experimental results were produced and validated using our implemented evaluation procedures rather than accepted directly from generative AI outputs. Figures and visualizations produced with AI assistance were manually inspected and revised to ensure that they accurately represented the underlying methodology and results. Synthetic trajectories generated by language models were subjected to the verification and filtering procedures described in the paper before being included in the training data. 

The authors made all final decisions regarding the methodology, experiments, analysis, interpretation, and presentation of the work. We take responsibility for the final content of this paper, including all text, claims, code, data, and artifacts produced with the aid of generative AI.

\subsection*{Ethics Statement}

This work does not involve human-subject experiments or the collection of new personally identifiable information. Our experiments use existing question-answering benchmarks and text corpora, and therefore inherit potential biases, inaccuracies, and other limitations present in these resources. The language models used to generate synthetic training trajectories may similarly reproduce biases or factual errors from their training data. To reduce the impact of erroneous synthetic supervision, our cold-start generation pipeline includes explicit execution-based verification and filtering before generated trajectories are used for training.

GrepSeek allows an agent to interact with a corpus through executable shell commands. Although our experiments restrict this interaction to a controlled corpus-search environment, more general deployment of executable agents may introduce security risks if commands are given unrestricted access to external systems, files, or sensitive data. Appropriate sandboxing, access control, command restrictions, and auditing should therefore be employed when adapting such systems to real-world environments. As with other retrieval and question-answering systems, downstream users should also consider the licensing, privacy, and usage requirements of the corpora on which the system is deployed. We have aimed to document the capabilities and limitations of the proposed approach, including cases in which lexical matching may fail because of ambiguity or surface-form variation.

\subsection*{Reproducibility Statement}

We provide the methodological and experimental details required to reproduce our results throughout the paper and appendices. Section~\ref{sec:method} describes the DCI agent, the cold-start data generation procedure, supervised fine-tuning, reinforcement learning with GRPO, and the optimized corpus-execution mechanisms. The corresponding algorithms, prompts, reward specification, and implementation details are provided in Appendix~\ref{app:our-implementation}. Section~\ref{sec:exp-setup} describes the datasets, corpus construction, evaluation metrics, baseline configurations, retrievers, training setup, and inference settings, with complete SFT, GRPO, and inference hyperparameters reported in the appendix. Additional evaluation results, ablations, and qualitative examples are provided in the remaining appendices.

\newpage

\bibliography{iclr2027_conference}
\bibliographystyle{iclr2027_conference_arxiv}

\appendix

\newpage
\section{Datasets}
\label{app:dataset}

Following prior work \citep{search-r1}, we evaluate our method on a comprehensive suite of eight knowledge-intensive benchmark datasets. These datasets are carefully selected to evaluate both single-step fact retrieval and complex, multi-step reasoning. To standardize the formatting and evaluation protocol, all datasets are obtained from the FlashRAG\footnote{Available at: \url{https://hf.co/datasets/RUC-NLPIR/FlashRAG_datasets}} and BrowseComp-Plus\footnote{Available at: \url{https://huggingface.co/datasets/Tevatron/browsecomp-plus}} repositories.

\subsection{Evaluation Benchmarks}

The evaluation suite is divided into single-hop and multi-hop datasets to isolate the agent's ability to perform targeted retrieval versus iterative corpus exploration.

\paragraph{Single-Hop Datasets} These tasks require retrieving a single, highly relevant fact or document to answer the user's query:
\begin{itemize}[leftmargin=1.2em]
    \item \textbf{Natural Questions (NQ) \citep{nq}:} A dataset of real user queries issued to Google Search. We use the open-domain split, which requires systems to retrieve relevant Wikipedia passages to answer questions formulated by users without prior knowledge of the target.
    \item \textbf{TriviaQA \citep{triviaqa}:} A collection of complex trivia questions authored by trivia enthusiasts. While the questions often contain compositional linguistic structures, the answers can typically be derived from a single retrieved document. 
    \item \textbf{PopQA \citep{popqa}:} An entity-centric QA dataset specifically designed to probe long-tail knowledge. The dataset is constructed from Wikidata triples and focuses on rare entities where parametric knowledge in LLMs typically fails, strictly necessitating accurate external retrieval.
\end{itemize}

\begin{table}[!t]
\caption{Dataset sizes for training and evaluation. Datasets marked with an asterisk ($^*$) indicate those utilized during the training phase for SFT and RL optimization.}
\label{tab:dataset}
\centering
\begin{tabular}{lrrr}
\toprule
 & \multicolumn{2}{c}{\textbf{Training Size}} & \\
\cmidrule(lr){2-3}
\textbf{Dataset} & \textbf{SFT} & \textbf{RL} & \textbf{Evaluation Size} \\
\midrule
NQ$^*$ & 5,000 & 79,168 & 3,610 \\
TriviaQA & -- & -- & 11,313 \\
PopQA & -- & -- & 14,267 \\
HotpotQA$^*$ & 5,000 & 90,447 & 7,405 \\
2WikiMultiHopQA & -- & -- & 12,576 \\
MuSiQue & -- & -- & 2,417 \\
Bamboogle & -- & -- & 125 \\
BrowseComp-Plus & -- & -- & 830 \\
\midrule
\textbf{Total} & \textbf{10,000} & \textbf{169,615} & \textbf{52,543} \\
\bottomrule
\end{tabular}
\end{table}

\paragraph{Multi-Hop Datasets} These tasks require the agent to execute multiple interdependent search queries, gathering partial information to inform subsequent retrieval steps:
\begin{itemize}[leftmargin=1.2em]
    \item \textbf{HotpotQA \citep{hotpotqa}:} A dataset requiring reasoning across at least two distinct Wikipedia articles. Questions are specifically designed to require information from multiple sources to synthesize and produce the correct final answer.
    \item \textbf{2WikiMultiHopQA (2Wiki) \citep{2wiki}:} Constructed using Wikidata properties to generate compositional questions. This dataset introduces explicit logical structures to the multi-hop reasoning process, requiring the agent to follow strict reasoning chains across multiple documents.
    \item \textbf{MuSiQue \citep{musique}:} A rigorous multi-hop QA dataset designed to minimize ``shortcut'' reasoning. The questions are composed by chaining multiple single-hop questions, heavily filtered to ensure that models cannot guess the answer through lexical overlap or single-document retrieval.
    \item \textbf{Bamboogle \citep{bamboogle}:} A smaller but highly challenging dataset consisting of questions manually authored to defeat standard search engines. It requires deep, multi-step evidence gathering that cannot be resolved using surface-level web snippets or simple entity linking.
    \item \textbf{BrowseComp-Plus \citep{chen2025browsecompplusfairtransparentevaluation}:} A controlled deep-search benchmark derived from BrowseComp, featuring challenging long-horizon, multi-hop questions over a fixed corpus. It requires iterative retrieval and reasoning across multiple pieces of evidence while enabling reproducible evaluation without live web search.
\end{itemize}

\begin{figure*}[!t]
\begin{promptbox}[DCI Agent System Prompt]
You are a research agent that searches a Wikipedia corpus to answer multi-hop questions by issuing shell commands.

Corpus file: corpus.jsonl

Treat the corpus as a large text file with one Wikipedia passage per line. The file has 21 million lines, so always cap output (use `| head -n 3` for narrow searches, up to `| head -n 8` when you need to scan more chunks of the same article).

Useful command patterns:
- Substring search:
    rg -F "distinctive phrase" corpus.jsonl | head -n 3
- AND-narrow with a second grep:
    rg -F "phrase1" corpus.jsonl | rg -i -F "phrase2" | head -n 3
- Count first, then narrow if too many results:
    rg -F "pattern" corpus.jsonl | wc -l
- Useful flags: -F (fixed string, no regex), -i (case-insensitive), -w (whole word).

Allowed shell tools: rg, grep, find, sed, awk, head, tail, cat, ls, wc, sort, cut, uniq, tr.
You MAY pipe with |. You MAY NOT use redirection (>, <), command chaining (; && ||), or command substitution (\$(...), `...`).

For every step, write a short paragraph of reasoning in plain prose (2-5 sentences) — what you have learned from prior tool outputs, what's still missing, what you'll search for next. Then output exactly one of:

<tool_call>
{"name": "shell", "arguments": {"command": "your single-pipeline shell command, no newlines"}}
</tool_call>

OR (only when you have enough information to answer):

<answer>
the final answer (concise — typically a short noun phrase, name, or date)
</answer>

Always reason first, then exactly one action block. Do not skip the reasoning.
\end{promptbox}
\vspace{-0.4cm}
\caption{System prompt for \ourmethod.}
\label{fig:sys-prompt}
\end{figure*}

For datasets from FlashRAG, we use the 2018 Wikipedia dump \citep{karpukhin2020dense} of 21M documents as the corpus.\footnote{Available at: \url{https://hf.co/datasets/PeterJinGo/wiki-18-corpus}} For BrowseComp-Plus, we use its accompanying corpus of 100,195 documents, chunked into 864,837 passages using 700-word windows with a stride of 630 words (70-word overlap).

\subsection{Data Splits and Training Protocol}
\label{app:data-splits}

To evaluate the generalization capabilities of our approach, we employ a strict split between in-distribution training datasets and out-of-distribution evaluation datasets.  For the training phase, the agent is trained exclusively on a combined dataset consisting of the training splits of \textbf{NQ} ($79,168$ examples) and \textbf{HotpotQA} ($90,447$ examples). This provides the agent with exposure to both fundamental single-hop retrieval dynamics and complex multi-hop reasoning strategies, totaling $169,615$ training examples. During inference, we evaluate the system across all eight datasets to test both held-out in-domain performance and generalization to unseen datasets. We utilize the official test splits for datasets where they are publicly available. The final evaluation encompasses $52,543$ total queries across the eight benchmarks. Full dataset sizes and split statistics are detailed in Table~\ref{tab:dataset}.

\subsection{Evaluation Metrics}
\label{sec:eval-metrics}

For datasets from FlashRAG, we evaluate generated answers using Exact Match (EM) and token-level F$_1$ \citep{rajpurkar-etal-2016-squad}. For BrowseComp-Plus, we follow its evaluation protocol and report LLM-judged accuracy using GPT-4.1,\footnote{\url{https://platform.openai.com/docs/models/gpt-4.1}} which determines whether the predicted final answer is correct given the reference answer.

\begin{itemize}[leftmargin=1.2em]
\item \textbf{Exact Match (EM):} The percentage of predictions that exactly match a gold answer after standard normalization, including lowercasing, punctuation removal, and removal of articles (e.g., `a'', `an'', ``the'').

\item \textbf{F$_1$ Score:} Token-level F$_1$ measures partial overlap between the prediction and reference answer after the same normalization. When multiple references are available, we report the maximum score across references.

\item \textbf{BrowseComp-Plus Accuracy:} Following the benchmark evaluation protocol, GPT-4.1 acts as an LLM judge to assess whether the predicted final answer matches the reference answer.

\end{itemize}

In the main paper, we report F$_1$ for the seven FlashRAG datasets and judge-based accuracy for BrowseComp-Plus; EM results for all eight datasets are provided in the appendix.

\begin{figure*}[!t]
\begin{promptbox}[Corpus description (shared by all roles)]
Corpus file: corpus.jsonl
Treat the corpus as a large text file with one Wikipedia passage per line. The file has 21 million lines, so always cap output (use `| head -n 3` for narrow searches, up to `| head -n 8` when you need to scan more chunks of the same article).
Useful command patterns:
- Substring search:
    rg -F "distinctive phrase" corpus.jsonl | head -n 3
- AND-narrow with a second grep:
    rg -F "phrase1" corpus.jsonl | rg -i -F "phrase2" | head -n 3
- Count first, then narrow if too many results:
    rg -F "pattern" corpus.jsonl | wc -l
- Useful flags: -F (fixed string, no regex), -i (case-insensitive), -w (whole word).
Allowed shell tools: rg, grep, find, sed, awk, head, tail, cat, ls, wc, sort, cut, uniq, tr.
You MAY pipe with |. You MAY NOT use redirection (>, <), command chaining (; && ||), or command substitution (\$(...), `...`).
\end{promptbox}
\vspace{-0.4cm}
\caption{System prompt describing the corpus and allowed shell tools.}
\label{fig:prompt_corpus_desc}
\end{figure*}

\newpage
\section{Detailed Related Work}
\label{app:detailed-related-work}

\paragraph{Retrieval-Augmented Agentic Search:}

Knowledge-intensive question answering is challenging for LLMs because many questions require facts that may be missing, outdated, or unreliable in the model's parametric memory \citep{popqa}. Access to external knowledge is central to improving factuality and coverage. Retrieval-Augmented Generation (RAG) \citep{rag} addresses this need by retrieving relevant evidence from an external corpus and conditioning generation on the retrieved context. More broadly, this can be viewed as an instance of retrieval-enhanced machine learning \citep{REML}. In this paradigm, the retrieval module determines what information is exposed to the model, typically through an index-based interface such as sparse lexical retrieval with BM25 \citep{bm25} or dense retrieval with embedding models such as E5 \citep{e5} and Qwen3-Embedding \citep{qwen3embedding}. This retrieval step gives the model access to non-parametric knowledge and has become a standard approach for knowledge-intensive tasks.

However, a single retrieval step is often insufficient for complex questions. Many information needs require intermediate reasoning: the system may need to identify an entity, use that entity to form a follow-up query, retrieve additional evidence, and then compose information across documents. This motivates retrieval-augmented reasoning methods, where retrieval is not merely a preprocessing step but part of a multi-turn reasoning process. Such multi-turn behavior is important for both unstructured corpora and structured settings, where systems may need to inspect intermediate results, refine constraints, and issue follow-up operations over databases or tables, as in STARQA \citep{starqa}. IRCoT \citep{ircot} interleaves chain-of-thought reasoning with retrieval, while more recent search-agent systems extend this idea to longer-horizon tool-use trajectories, including Search-R1 \citep{search-r1} and Search-O1 \citep{search-o1}.

Recent agentic search methods differ in which parts of the pipeline are optimized. Some systems treat both the LLM and the retriever or search engine as black boxes, relying on prompting or inference-time orchestration to decide when and how to search, as in Search-O1 \citep{search-o1}. Other methods train the language model to issue better search queries and reason over retrieved evidence while keeping the underlying retriever fixed, as in Search-R1 \citep{search-r1}. A third line of work jointly optimizes the reasoning policy and the retrieval or ranking component, as in CoSearch \citep{zeng2026cosearch}. These approaches primarily differ in how they improve reasoning, planning, or retrieval over a conventional search interface. Our work studies a complementary direction: instead of training an agent to better use a fixed retriever, we train a compact open-weight model to interact directly with the corpus through deterministic shell-based search operations.

Question answering is widely used to evaluate retrieval-augmented and agentic search systems because it directly tests whether a system can retrieve sufficient evidence and synthesize the correct answer. Multi-hop QA benchmarks are especially relevant because they require multi-turn search behavior: a model must often retrieve one piece of evidence, use it to identify a new information need, and then combine evidence across steps. Following prior search-agent work such as Search-R1 \citep{search-r1}, we evaluate on both single-hop and multi-hop QA benchmarks. 


\paragraph{Direct Interaction with Corpus:} 

Direct Corpus Interaction (DCI) provides a different way to connect language models with external information. Instead of relying on a retriever to rank passages, the agent issues explicit operations over the raw corpus and controls how evidence is matched, filtered, and composed. Prior work has studied direct textual search in code and repository settings \citep{10.1145/3565971,wang2026grepragempiricalstudyoptimization}, as well as recent DCI-style agents for open-domain retrieval over large-scale corpora \citep{sen2026grepneedagentharnesses,dci-agent, subramanian2025keywordsearchneedachieving}. This direction is also related to systems work on efficient string and regular-expression search over large or compressed text collections, including Succinct and Swift \citep{188994,NAVARRO2003423}. We view these systems primarily as efficiency-oriented predecessors rather than agentic-search baselines: they improve the execution substrate for direct search, while our focus is on whether an LLM can learn when and how to use direct corpus operations as part of multi-step reasoning.

\newpage
\section{\ourmethod's Implementation Details}
\label{app:our-implementation}

This section provides implementation details and settings used for \ourmethod.

\subsection{Prompts}

The main system prompt for the DCI agent is shown in Figure~\ref{fig:sys-prompt}. The prompts described in this section collectively define the instructional framework used to generate synthetic cold-start training trajectories. All agent roles share a common system prompt that specifies the corpus structure, interaction format, and permissible shell tools (Figure~\ref{fig:prompt_corpus_desc}).


\begin{figure*}[!b]
\begin{promptbox}[Decomposition]
You are decomposing a multi-hop question into the ordered chain of single-hop sub-questions an agent would need to solve in order to reach the answer.

Question: {question}
Final answer: {answer}

Rules:
- Output 1 to 3 ordered sub-questions; each one's answer is what an agent would need before it can solve the next one.
- Order MUST match the order an agent would solve the question (sub_q[i+1] depends on sub_q[i]'s answer).
- The LAST sub-question's answer is the final answer above.
- Do NOT include expected_answers — only the sub-questions themselves. The intermediate answers will be derived later.
- Do NOT make a sub-question whose answer is already given in the question (e.g., do not ask "Who is X?" when the question already names X).

Output ONLY a JSON array, no prose:
[{{"sub_question": "..."}}, {{"sub_question": "..."}}, ...]
\end{promptbox}
\vspace{-0.4cm}
\caption{Tutor prompt for decomposing multi-hop questions into single-hop steps.}
\label{fig:prompt_decomposition}
\end{figure*}

In \textbf{Phase A}, the pipeline employs an Answer-Aware Tutor to decompose multi-hop questions into ordered single-hop sub-queries (prompt in Figure~\ref{fig:prompt_decomposition}). Guided by system instructions that enforce the anti-leak constraint (prompt in Figure~\ref{fig:prompt_backward_system}), the Tutor iteratively constructs a backward retrieval trajectory through an initial command proposal stage (prompt in Figure~\ref{fig:prompt_backward_first}) followed by targeted refinement steps (prompt in Figure~\ref{fig:prompt_backward_refine}). Each retrieved document is validated using a per-hop entailment judge (Figure~\ref{fig:prompt_judge}), after which a dedicated bridge extraction prompt identifies the intermediate entity required to connect the current retrieval step to the preceding sub-query (prompt in Figure~\ref{fig:prompt_bridge_extraction}).

\begin{figure*}[!t]
\begin{promptbox}[Per-hop judge (does the doc confirm the answer?)]
You are judging whether a tool output confirms the expected answer to a sub-question.
Sub-question: {sub_question}
Expected answer (primary): {expected_answer}
Acceptable alternative forms (any of these counts as confirming): {acceptable_forms}
Tool output:{tool_output}
Read the tool output. Does it contain a passage that clearly answers the sub-question with the primary expected answer OR any of the acceptable alternative forms? Be strict — a tangential mention or a different entity with a coincidentally similar name does NOT count. But if the same underlying entity is referenced under any of the alternative forms (e.g., a stage name, a real name, or an alias listed above), that DOES count as confirming.
Output exactly one JSON object:
{{"verdict": "YES" or "NO", "reasoning": "<one short sentence>"}}
\end{promptbox}
\vspace{-0.4cm}
\caption{Prompt for judging if the retrieved document entails the target answer.}
\label{fig:prompt_judge}
\end{figure*}

\begin{figure*}[!t]
\begin{promptbox}[Backward command -- refine attempt]
Your previous command did not retrieve a passage that confirms the expected answer.
Sub-question: {sub_question}
Expected answer (do NOT search for this string): {expected_answer}
Alternative forms (also forbidden as search terms):{forbidden_forms}
Previous attempts:{prior_attempts}
Most recent attempt: Command: {last_command} Output:{last_output}
Judge said it does NOT confirm the answer because:{judge_reasoning}
Propose a different command. Common fixes:
- Your phrase may not appear verbatim — try a shorter or differently-worded substring from the sub-question.
- The query may be too broad — add a second piped rg with another distinctive phrase to AND-narrow.
- The query may be too narrow — drop a filter or shorten the phrase to broaden.
- Try synonym variants of words from the sub-question (e.g., "headquartered in" vs "head office", "based in", "located in").
Same hard rules as before (do NOT use the expected answer as a search term; keep output <= 5 lines via head; one pipeline; whitelisted tools only).
Output exactly:<reasoning>why the previous command failed and what is different about your new approach</reasoning><command>
your new single-pipeline shell command</command>
\end{promptbox}
\vspace{-0.4cm}
\caption{Tutor prompt for refining a failed retrieval command.}
\label{fig:prompt_backward_refine}
\end{figure*}

After constructing a verified retrieval path, \textbf{Phase B} transitions to an Answer-Blind Planner operating under a separate forward-acting system prompt (prompt in Figure~\ref{fig:prompt_planner_system}). At each step, the Planner generates an initial history-conditioned reasoning trace and retrieval action proposal (prompt in Figure~\ref{fig:prompt_planner_step}). A Tutor-edit stage then refines the Planner's reasoning to align it with the verified retrieval command while remaining strictly grounded in the observable interaction history (prompt in Figure~\ref{fig:prompt_tutor_edit}). This process produces trajectories that preserve realistic forward causal reasoning while avoiding future-state leakage introduced by backward evidence construction.

\begin{figure*}[!t]
\begin{promptbox}[Bridge-entity extraction]
You are reading a passage that was retrieved to help answer a multi-hop question. Your job is to identify the entity the passage uses as the bridge — the answer to the previous sub-question.
Original multi-hop question: {question}
Earlier sub-question (whose answer we are identifying): {sub_q_prev}
Later sub-question (already solved): {sub_q_next}
Answer to the later sub-question: {expected_next}
Passage retrieved for the later sub-question:{doc_next}
Read the passage. The passage answers the later sub-question; somewhere in it, the entity named in the answer to the EARLIER sub-question appears (since the later sub-question depends on it). Extract that entity exactly.
Also collect:
- "aliases" — alternative names of the SAME entity that the passage explicitly mentions (e.g., stage names, real names, abbreviations, parenthetical "also known as"/"born ..." forms). Only include forms that literally appear in the passage. Do NOT invent aliases.
- "alternates" — at most 2 OTHER plausible candidate entities from the passage that could also be the bridge if your primary pick is wrong. Use this only when the passage genuinely supports multiple readings.
Output a JSON object, nothing else: {{"bridge_entity": "<short noun phrase>", "aliases": ["<alt-name-1>", ...], "alternates": ["<other-candidate-1>", ...], "evidence": "<the exact short snippet from the passage that names the primary entity>"}}
If you cannot extract a clear bridge entity (the passage truly does not contain enough to derive the previous answer), output:
{{"bridge_entity": null, "aliases": [], "alternates": [], "evidence": null}}
\end{promptbox}
\vspace{-0.4cm}
\caption{Prompt for extracting the bridging entity required for backward chaining.}
\label{fig:prompt_bridge_extraction}
\end{figure*}


\begin{figure*}[!t]
\begin{promptbox}[Planner -- system (answer-blind agent)]
You are a research agent that searches a Wikipedia corpus to answer multi-hop questions by issuing shell commands.
{corpus_description}
For every step, write a short paragraph of reasoning in plain prose (2-5 sentences) — what you have learned from prior tool outputs, what's still missing, what you'll search for next. Then output exactly one of:
<tool_call>
{{"name": "shell", "arguments": {{"command": "your single-pipeline shell command, no newlines"}}}}
</tool_call>
OR (only when you have enough information to answer):
<answer>
the final answer (concise — typically a short noun phrase, name, or date)
</answer>
Always reason first, then exactly one action block. Do not skip the reasoning.
\end{promptbox}
\vspace{-0.4cm}
\caption{System prompt for the Planner agent used during forward assembly.}
\label{fig:prompt_planner_system}
\end{figure*}

\begin{figure*}[!t]
\begin{promptbox}[Planner -- step user prompt]
Question: {question}

Trace so far:
{history}

Produce the next step.
\end{promptbox}
\vspace{-0.4cm}
\caption{Standard user prompt for the Planner agent during trajectory generation.}
\label{fig:prompt_planner_step}
\end{figure*}

\begin{figure*}[!t]
\begin{promptbox}[Final answer user prompt]
Question: {question}

Trace so far:
{history}

You now have enough information. Produce a brief reasoning paragraph synthesizing the answer from the trace, then output exactly:

<answer>
the final answer (concise — just a name, date, or short noun phrase)
</answer>
\end{promptbox}
\vspace{-0.4cm}
\caption{User prompt for the final answer formulation step.}
\label{fig:prompt_final_answer}
\end{figure*}

\begin{figure*}[!t]
\begin{promptbox}[Tutor edit (rewrite think to reach target command)]
You are editing a research agent's draft reasoning so that it leads naturally to a known-correct next command. Your goal is to produce a final reasoning paragraph that: 1. Is grounded ONLY in what the agent has actually observed so far (the trace below). Do NOT add facts the agent could not know at this point. 2. Concludes with a clear motivation for the target command. 3. Reads as natural forward reasoning by an agent — not as a justification written after the fact. Hedge where genuine uncertainty exists. 4. Stays close to the agent's draft when draft is already on track; rewrite freely when the draft proposes something the wrong direction.
Original question (the agent knows this): {question}
Trace observed so far by the agent: {history}
Agent's draft reasoning: {draft_think}
Agent's draft command (may be discarded): {draft_command}
Known-correct next command (what trace will use): {target_command}
What this command will return when executed (for your context only — the agent only sees this AFTER it runs the command, NOT while reasoning): {target_doc_preview}
FORBIDDEN — your edited reasoning must NOT contain: (a) expected answer of current step, final answer, or any answer to be discovered in a future step. agent should not know these yet. (b) Any factual claim agent could not derive from (i) user's question, (ii) prior tool_responses in trace, or (iii) common knowledge.
ALLOWED — the agent's reasoning may freely include: - Words from the question and synonyms/paraphrases of them ("hotel" -> "hospitality","head office"->"headquarters"/"based in"). - Tentative hypotheses about WHERE answer might be found, phrased as guesses — e.g., "article likely uses phrase 'headquartered in' rather than 'head office'", or "this is probably described in a Wikipedia article about family's business". - Common-knowledge inference — e.g., "Delhi is the capital of India" once Delhi has appeared, or "World Cup 2002 was the FIFA tournament" given a sports question.
Concrete pass/fail examples: Question: "The Oberoi family is part of a hotel company that has a head office in what city?" Trace so far: (no commands run yet). Target command: rg -F 'Oberoi family' corpus.jsonl | rg -F 'hotel' | head -n 3
LEAK (names bridge answer "Oberoi Group" before it has been seen): "I need to locate article on Oberoi Group to find its headquarters."
OK (describes the search in question terms with hypotheses): "I need to find which hotel company the Oberoi family is part of. The corpus likely contains passages mentioning the 'Oberoi family' alongside hotel-business context. I'll grep for 'Oberoi family' and narrow with 'hotel' to surface the relevant article."
Question: same as above. Trace so far: tool_response showed "...the Oberoi family is the majority shareholder in EIH Ltd, parent of The Oberoi Group..." Target command: rg -F 'Oberoi Group' corpus.jsonl | rg -F 'head office' | head -n 3
OK ("Oberoi Group" is now derivable from previous tool_response): "The previous result identifies family's hotel company as The Oberoi Group. To find its head office, I'll search for that company alongside 'head office', or a synonym like 'headquartered'."
When planner's draft is already valid under these rules, keep it close to draft. When draft proposes a substantively different direction, rewrite the reasoning so it naturally leads to the target command — using hypotheses, synonyms, and search-strategy reasoning, never by naming entities the agent shouldn't know yet.
Output:<edited_reasoning>final reasoning</edited_reasoning>
\end{promptbox}
\vspace{-0.4cm}
\caption{Tutor prompt for steering agent reasoning toward verified actions.}
\label{fig:prompt_tutor_edit}
\end{figure*}

Finally, in \textbf{Phase C}, the Planner generates a final answer conditioned exclusively on the accumulated interaction trajectory (prompt in Figure~\ref{fig:prompt_final_answer}). The completed trajectory is then evaluated by a Trajectory Coherence Judge (prompt in Figure~\ref{fig:prompt_coherence_judge}), which enforces strict temporal consistency constraints and rejects trajectories that implicitly reveal target entities, retrieval terms, or unobserved facts before they become available within the agent's causal history.


\begin{figure*}[!t]
\begin{promptbox}[Trajectory coherence judge]
You are an expert reviewer of multi-hop QA trajectories produced by a research agent. The agent answers a question by issuing shell commands against a corpus and reasoning about the results across multiple turns. Your job is NOT to solve the question. Your job is to verify that the trajectory is INTERNALLY COHERENT.
INFORMATION FRONTIER
At any TURN k, the agent could legitimately know only:
  (a) the text of the original QUESTION,
  (b) anything appeared in the OUTPUT of any earlier turn (1 ... k-1),
  (c) generic common knowledge that any educated reader would have without consulting Wikipedia. Examples of (c):
    - "Delhi in India", "Brazil is country", "WW II ended in 1945".
    - "Headquarters"/"head office"/"based in" mean roughly the same.
    - Synonyms and paraphrases of words from the question.
      Counter-examples (NOT (c)):
    - "Cafu was Brazil's 2002 World Cup captain" - too specific.
    - "The Oberoi Group is based in Delhi" - too specific.
    - The bridge answer of any sub-question.
A turn explicitly tagged "[self-correction: failed attempt by design]" is allowed to issue a wrong command and produce confident reasoning that leads to that wrong command. A turn tagged "[recovery from TURN k]" is allowed to reference the failed turn's OUTPUT as observed evidence.

CHECKS: The trajectory FAILS if ANY of the following holds for ANY turn. Report the FIRST turn where it fails and which check failed.
CHECK 1 - REASONING leaks future facts. TURN k's REASONING names a specific entity, number, date, or fact not derivable from the information frontier (excluding self-correction turns). FAIL example: TURN 1 REASONING (no prior turns) says "I'll look up the Oberoi Group's headquarters" when the question only mentions "Oberoi family" - the company name has not been observed.
CHECK 2 - COMMAND leaks future facts. TURN k's COMMAND uses a search term not derivable from the information frontier. FAIL example: question asks for a track length, no prior OUTPUT has named a number, and the COMMAND contains rg -F "6.213" - that number is the answer the agent should still be searching for.
CHECK 3 - ACTION does not match REASONING. TURN k's REASONING explicitly states "the answer is X", "no further search is needed", "the question is answered", "I have the answer", and the action on this turn is a COMMAND rather than a FINAL ANSWER. FAIL example: REASONING ends with "the answer is clearly Amy Jo Johnson, no further tool execution is needed" and the action is a shell command.
CHECK 4 - FINAL ANSWER not supported. The final turn's FINAL ANSWER is not stated in, or clearly inferrable from, any earlier OUTPUT. FAIL example: FINAL ANSWER is "Roseau, Minnesota" but no earlier OUTPUT contains "Roseau" or supports that location.
If none of CHECK 1-4 fails on any turn, the trajectory PASSES.
When in doubt, prefer FAIL over PASS - we want clean training data, not high yield.
OUTPUT: Return EXACTLY one JSON object, no surrounding prose, no markdown fences:{{"verdict": "PASS" or "FAIL", "failing_check": 1 or 2 or 3 or 4 or null, "first_failing_turn": <int or null>, "reasoning": "<one short sentence, max ~30 words>"}}
QUESTION: {question}
TRAJECTORY: {trajectory_text}
\end{promptbox}
\vspace{-0.4cm}
\caption{Prompt for the final quality gate checking for information leakage.}
\label{fig:prompt_coherence_judge}
\end{figure*}

\begin{algorithm*}[!t]
\caption{Cold-start Trajectory Generation for \ourmethod{}}
\label{alg:coldstart}
\begin{algorithmic}[1]
\Require query $q$, gold answer $y$; corpus $\mathcal{C}$; max refinement iterations $M$; tutor $\mathcal{M}_{T}$, planner $\mathcal{M}_{P}$
\Ensure  Verified training trajectory $\mathcal{T}_{\mathrm{train}}$, or \textsc{Fail}

\Statex \textbf{\textit{// Phase A: Goal-Aware Decomposition and Backward Verification}}
\State $(q_1,\dots,q_N) \gets \mathcal{M}_{T}.$\textsc{Decompose}($q, y$) \Comment{Generate ordered sub-queries; \textsc{Fail} if unparsable} \label{line:decompose}
\State $a \gets y$;\ \ $F \gets \emptyset$;\ \ $\mathcal{D} \gets \emptyset$;\ \ $\mathcal{S} \gets \langle \, \rangle$ \Comment{$a$: target entity; $F$: aliases; $\mathcal{D}$: evidence context} \label{line:init_vars}
\For{$i = N$ \textbf{down to} $1$} \label{line:backward_loop}
    \State $(\textit{success}, c_i, d_i) \gets$ \textsc{Discover}($q_i, a, F, \mathcal{D}$);\ \ \textbf{if} $\neg\,\textit{success}$ \textbf{then return} \textsc{Fail} \label{line:discover}
    \State $\mathcal{S} \gets \mathcal{S} \oplus \langle i, q_i, a, c_i, d_i \rangle$;\ \ $\mathcal{D} \gets \mathcal{D} \cup \{d_i\}$ \label{line:record_step}
    \If{$i > 1$} \label{line:if_bridge}
        \State $(a, F) \gets$ \textsc{GetBridge}($q, q_{i-1}, q_i, a, d_i$);\ \ \textbf{if} $a = \emptyset$ \textbf{then return} \textsc{Fail} \label{line:bridge}
    \EndIf
\EndFor

\Statex \textbf{\textit{// Phase B: Forward Assembly (Answer-Blind Planner, Tutor-Guided)}}
\State $\mathcal{S} \gets$ \textsc{Reverse}($\mathcal{S}$);\ \ $\mathcal{H} \gets \langle \, \rangle$ \Comment{Restore forward order; init state history} \label{line:reverse_init}
\For{\textbf{each} $\langle i, q_i, a, c_i, d_i \rangle \in \mathcal{S}$} \label{line:forward_loop}
    \State $(\theta_d, c_d) \gets \mathcal{M}_{P}.$\textsc{Draft}($q, \mathcal{H}$) \Comment{Answer-blind prediction of reasoning and action} \label{line:draft}
    \State $\theta \gets \mathcal{M}_{T}.$\textsc{Align}($q, \mathcal{H}, \theta_d, c_d, c_i, d_i$);\ \ \textbf{if} $\theta = \emptyset$ \textbf{then return} \textsc{Fail} \label{line:align}
    \State $\mathcal{H} \gets \mathcal{H} \oplus \langle \theta, c_i, d_i \rangle$ \Comment{Ensure reasoning is strictly conditioned on causal history} \label{line:append_h}
\EndFor

\Statex \textbf{\textit{// Phase C: Answer Formulation and Quality Assurance}}
\State $\hat{y} \gets \mathcal{M}_{P}.$\textsc{Answer}($q, \mathcal{H}$) \label{line:answer}
\State $\mathcal{T}_{\mathrm{train}} \gets$ \textsc{Format}($q, \mathcal{H}, \hat{y}$) \label{line:format}
\State \textbf{if} $\hat{y} = \emptyset \lor \mathrm{F_1}(\hat{y}, y) = 0 \lor$ \textsc{Judge}($q, \mathcal{T}_{\mathrm{train}}$) $\neq \mathrm{Pass}$ \textbf{then return} \textsc{Fail} \label{line:judge_gate}
\State \Return $\mathcal{T}_{\mathrm{train}}$ \label{line:return_success}

\Statex
\Function{Discover}{$q', a, F, \mathcal{D}$} \Comment{Identify a target command that retrieves evidence for $a$} \label{line:func_discover}
    \For{$t = 1, \dots, M$} \label{line:refine_loop}
        \State $c \gets \mathcal{M}_{T}.$\textsc{Propose}($q', a, \{a\} \cup F, \mathcal{D}$) \Comment{Strictly exclude target \& aliases} \label{line:propose}
        \State $d \gets$ \textsc{Exec}($c, \mathcal{C}$);\ \ \textbf{if} $\mathcal{M}_{T}.$\textsc{Check}($q', a, d, F$) $= \mathrm{True}$ \textbf{then return} $(\mathrm{True}, c, d)$ \label{line:exec_check}
    \EndFor
    \State \Return $(\mathrm{False}, \emptyset, \emptyset)$ \label{line:return_fail}
\EndFunction
\end{algorithmic}
\end{algorithm*}

\subsection{Pruned Adaptive Command Execution (PACE)}
\label{app:eff-search-Pruned}

PACE accelerates DCI by reducing the corpus scanned for each command while preserving correctness. It conservatively identifies a superset of candidate lines that may match, then executes the original command only on those candidates. Thus, pruning may introduce false positives but never removes true matches. Commands that cannot be safely or efficiently pruned fall back to execution over the full corpus. We next describe the data structures enabling this pruning.

\subsubsection{Offline Auxiliary Structure Construction}

Pruned Adaptive Command Execution (PACE) constructs a compact auxiliary structure once for each corpus. This structure is used only to identify candidate document identifiers; the original corpus remains unchanged and is consulted during command execution.

\paragraph{Term Catalog and Posting Lists:}

The corpus is processed document-by-document. Each document is case-folded and decomposed into maximal Unicode alphanumeric runs, and terms shorter than two characters are omitted. For each distinct term $v$ in the corpus, we record a posting list $P(v)$ containing the identifiers of documents in which the term occurs. This is inspired by inverted indices in retrieval systems such as BM25. Since documents are processed in corpus order and each term is recorded at most once per document, the posting lists are naturally sorted and duplicate-free.

\paragraph{Substring-Searchable Vocabulary:}

Exact vocabulary lookup is insufficient for shell-based search because \texttt{rg} and \texttt{grep} perform substring matching. For example, a search for \texttt{16} must also consider vocabulary terms such as \texttt{2016} and \texttt{b16}.\footnote{This is necessary because \texttt{rg} and \texttt{grep} also support substring matching.} To support such lookups efficiently, we first serialize the vocabulary into a single byte sequence. Specifically, given vocabulary terms $v_1,\ldots,v_m$, we construct:
$
S = v_1 \,\Vert\, \texttt{\textbackslash0} \,\Vert\, v_2 \,\Vert\, \texttt{\textbackslash0} \,\Vert\, \cdots \,\Vert\, v_m \,\Vert\, \texttt{\textbackslash0},
$
where \texttt{\textbackslash0} is a separator that does not occur inside a vocabulary term. We also retain a mapping from each position in $S$ to the vocabulary term containing that position. We then construct a suffix array \citep{suffix-array} over $S$. Each suffix-array entry therefore corresponds to a starting position within one of the vocabulary terms. Because suffixes are sorted lexicographically, all suffixes beginning with the same byte sequence are contiguous. At query time, for a query term $q$, two binary searches locate the interval of suffixes having $q$ as a prefix. Since a suffix can start at any character position within a vocabulary term, this interval identifies exactly the vocabulary terms that contain $q$ as a substring. The corresponding term identifiers can then be mapped to their posting lists. We further compact the suffix array by discarding entries that can never match a valid query. PACE considers only query terms of length at least two. Consequently, a suffix beginning at a separator cannot match any query, and a suffix beginning at the final character of a vocabulary term contains fewer than two term characters before reaching the separator and therefore cannot match either. These suffixes are removed after construction while preserving the relative lexicographic order of all retained entries. This reduces the suffix-array size by approximately $22\%$ without changing the result of any valid substring lookup.

\paragraph{Compact Auxiliary Representation.}
PACE uses compact representations for each component needed to translate a substring match into candidate documents. First, a suffix-array lookup returns positions in the concatenated vocabulary rather than vocabulary-term identifiers. Storing an explicit term identifier for every suffix would require substantial additional memory. Instead, PACE stores a bitvector marking the null separators between vocabulary terms. For a suffix starting at byte position $p$, the containing term is identified by a rank query over this bitvector, which counts the number of separators preceding $p$. This reduces the suffix-to-term mapping from 932\,MB to 44\,MB on wiki18.

The resulting term identifiers are then mapped to their document posting lists. These lists are highly skewed: on wiki18, $88\%$ of the 26.7M vocabulary terms occur in exactly one document. PACE therefore represents singleton terms directly with a single 32-bit document identifier, avoiding the metadata required for a general posting list. Posting lists containing multiple document identifiers are stored using Elias--Fano encoding \citep{fano1971number, 10.1145/321812.321820, 10.1145/2433396.2433409}. For a sorted list of $n$ document identifiers drawn from a corpus of $u$ documents, Elias--Fano requires approximately
$2 + \log_2(u/n)$
bits per posting while supporting efficient sequential decoding. Together, singleton specialization and Elias--Fano compression reduce posting storage from 6.11\,GB to 1.68\,GB on wiki18. Once candidate identifiers are recovered, PACE locates the documents in the original JSONL corpus. Because each document occupies one line, PACE records its starting byte offset. Rather than storing every offset as a 64-bit integer, documents are partitioned into blocks of 1,024: each block stores one 64-bit base offset, while individual documents store a 32-bit delta relative to that base. This representation allows candidate documents to be accessed directly without scanning the corpus. Across all components, the resulting auxiliary structure occupies 2.79\,GB for the 14.39\,GB wiki18 corpus and 0.44\,GB for the 3.73\,GB BrowseComp-Plus corpus. All components are memory-mapped read-only, allowing concurrent execution workers to share the same physical memory pages.

\subsubsection{Online Cost-Aware Candidate Construction:}

During inference, PACE analyzes the emitted shell pipeline to determine whether candidate pruning can be applied safely (Algorithm~\ref{alg:pace}, line~\ref{line:pace-classify}). The first stage must be a positive \texttt{rg} or \texttt{grep} filter over the corpus. Subsequent stages are inspected while they remain document-wise filtering operations. Positive literal filters contribute constraints to candidate construction, whereas inverted filters are excluded because they do not imply that a matching document must contain a particular literal. Pipelines containing unsupported or globally stateful operations are routed directly to the original sequential execution path (line~\ref{line:pace-fallback}).

For each contributing literal, PACE applies the same case folding and tokenization used during offline construction (line~\ref{line:pace-tokenize}). For example, the literal \texttt{Ring-Road} produces the query terms \texttt{ring} and \texttt{road}. For each query term $q$, two binary searches over the suffix array identify the vocabulary terms containing $q$ as a substring (line~\ref{line:pace-suffix})~\citep{suffix-array}. The corresponding suffix positions are mapped to vocabulary-term identifiers using the separator-rank structure (line~\ref{line:pace-rank}). Their posting lists define the candidate set
$C(q) = \bigcup_{v \in V(q)} P(v)$, 
where $V(q)$ denotes the set of vocabulary terms containing $q$. The union is necessary because a substring match may occur within any such vocabulary term. When multiple positive query terms are available, their candidate sets can be intersected (line~\ref{line:pace-intersect}):
$C=\bigcap_{q \in Q'} C(q)$, where $Q' \subseteq Q$ contains the terms selected by PACE. This construction is conservative: every true matching document must occur in the posting-derived candidate set of every selected positive term. Consequently, omitting a term from $Q'$ can only enlarge $C$ and cannot remove a true match.

PACE does not necessarily materialize every available posting list. Once the suffix-array range for $q$ has been resolved, stored posting-list cardinalities provide the estimate
$\operatorname{est}(q)=\sum_{v \in V(q)} |P(v)|$, which upper-bounds $|C(q)|$ without decoding the postings themselves (line~\ref{line:pace-est}). Query terms are considered in increasing order of $\operatorname{est}(q)$ so that selective constraints can reduce the candidate set before less selective ones are processed (line~\ref{line:pace-sort}). The posting union of the most selective term initializes the candidate set (line~\ref{line:pace-initial-candidates}). After each intersection, PACE compares the expected cost of processing another term with the cost of directly retrieving and verifying the current candidates. This process stops when the current candidate set is sufficiently small, the next term is too frequent to provide useful pruning, or further posting-list processing is estimated to be more expensive than candidate verification (line~\ref{line:pace-stop}). If a query term has an empty suffix-array range, no vocabulary term contains it and the command can immediately be determined to have no matching document (line~\ref{line:pace-empty}). At the opposite extreme, highly frequent substrings may match very large portions of the vocabulary and are skipped when expanding them would provide little benefit (line~\ref{line:pace-sa-skip}). These decisions affect only execution cost: skipping a pruning constraint enlarges the candidate superset but does not affect correctness.

Candidate sets are represented according to their density. Sparse sets are maintained as sorted arrays of document identifiers, enabling efficient ordered intersections, whereas dense sets are represented as per-document masks so that unions and intersections can be performed using scatter and bitwise operations. Finally, PACE abandons candidate pruning when even the most selective available term is too frequent or when the resulting candidate set exceeds the configured document or byte budget (line~\ref{line:pace-budget}). In these cases, the command is executed through the original full-corpus path.

\subsubsection{Semantics-Preserving Execution}

Once a candidate set has been constructed, PACE retrieves the corresponding documents directly from the original corpus (Algorithm~\ref{alg:pace}, line~\ref{line:pace-materialize}). Each document identifier is translated to its byte position using the stored document-offset structure. Because each JSONL document occupies exactly one line, consecutive candidate identifiers correspond to adjacent regions of the corpus. PACE coalesces such candidates into contiguous byte ranges and reads these ranges from a memory-mapped view of the original file in increasing corpus order. Maintaining corpus order is necessary for downstream operations whose result depends on input order, such as \texttt{head -n N}. Candidate documents are therefore streamed in their original order and in bounded chunks rather than assembled into a single intermediate buffer. This also preserves the streaming behavior of standard shell pipelines: downstream stages can terminate after consuming sufficient input without requiring the remaining candidate documents to be read or processed.

The original command semantics are then applied to the materialized candidate documents (line~\ref{line:pace-exec}). Candidate pruning therefore changes only which corpus documents must be examined. The original filtering predicates are still applied before a document can contribute to the command output, while corpus ordering, output formatting, truncation, counting, and exit-status semantics are preserved.

\begin{algorithm}[t]
\caption{Pruned Adaptive Command Execution (PACE)}
\label{alg:pace}
\begin{algorithmic}[1]

\Require Command $c$; corpus $\mathcal{C}$ with $u$ documents; vocabulary $\mathcal{V}$; suffix array $\mathrm{SA}$; separator-rank structure $\mathrm{Rank}$; singleton postings $\mathrm{Single}$; compressed postings $\mathrm{EF}$; posting counts $\mathrm{Count}$; document offsets $\mathrm{Off}$; hyperparameters $B_{\mathrm{SA}}$ (maximum suffix-array range), $B_{\mathrm{post}}$ (maximum posting estimate), $B_{\mathrm{target}}$ (target candidate count), $B_{\mathrm{doc}}$ (maximum candidate documents), and $B_{\mathrm{byte}}$ (maximum candidate bytes)

\Ensure Output identical to sequential execution of $c$ on $\mathcal{C}$

\State $(\tau,q) \gets \textsc{Classify}(c)$
    \label{line:pace-classify}
    \Comment{Check if can be executed fast and collect literal in command}

\State \textbf{if } $\tau=\textsc{Sequential}$ \textbf{ then } 
    \Return $\textsc{Exec}(c,\mathcal{C})$
    \label{line:pace-fallback}
    \Comment{Fallback for unsupported commands}

\State $Q \gets \textsc{Tokenize}(q)$
    \label{line:pace-tokenize}
    \Comment{Case-fold and extract query terms}
\State $\mathcal{R} \gets \emptyset$

\For{$q \in Q$}
    \State $[l_q,r_q) \gets \textsc{SuffixRange}(\mathcal{V},\mathrm{SA},q)$
        \label{line:pace-suffix}
        \Comment{Find vocabulary terms containing $q$}
    \State \textbf{if } $l_q=r_q$ \textbf{ then } 
        \Return $\textsc{ExecEmpty}(c)$
        \label{line:pace-empty}
        \Comment{No document can match}
    \State \textbf{if } $r_q-l_q>B_{\mathrm{SA}}$ \textbf{ then continue}
        \label{line:pace-sa-skip}
        \Comment{Skip overly common substring}

    \State $V(q)\gets
        \textsc{Unique}(\textsc{Rank}(\mathrm{SA}[l_q:r_q]))$
        \label{line:pace-rank}
        \Comment{Map suffixes to vocabulary terms}
    \State $e_q\gets\sum_{v\in V(q)} |P(v)|$
        \label{line:pace-est}
        \Comment{Estimate posting cost without decoding}
    \State $\mathcal{R}\gets\mathcal{R}\cup\{(V(q),e_q)\}$
\EndFor

\State \textbf{if } $\mathcal{R}=\emptyset$ \textbf{ then } \Return $\textsc{Exec}(c,\mathcal{C})$

\State $\mathcal{R}\gets\textsc{SortAscending}(\mathcal{R},e)$
    \label{line:pace-sort}
    \Comment{Most selective term first}
\State $(V,e)\gets\textsc{First}(\mathcal{R})$

\State \textbf{if } $e>B_{\mathrm{post}}$ \textbf{ then } \Return $\textsc{Exec}(c,\mathcal{C})$ \Comment{Even the cheapest term is too common}

\State $D\gets\bigcup_{v\in V}P(v)$
    \label{line:pace-initial-candidates}
    \Comment{Initial candidate documents}

\For{$(V,e)\in\mathcal{R}[2:]$}
    \If{$|D|\leq B_{\mathrm{target}}
        \ \lor\ e>B_{\mathrm{post}}
        \ \lor\ \textsc{GatherCost}(e)>\textsc{VerifyCost}(D)$}
        \State \textbf{break}
        \label{line:pace-stop}
        \Comment{Further pruning is not worthwhile}
    \EndIf
    \State $D\gets D\cap\left(\bigcup_{v\in V}P(v)\right)$
        \label{line:pace-intersect}
        \Comment{Intersect candidates across query terms}
    \State \textbf{if } $D=\emptyset$ \textbf{ then } \Return $\textsc{ExecEmpty}(c)$
\EndFor

\State \textbf{if } $|D|>B_{\mathrm{doc}}
    \lor \textsc{Bytes}(D,\mathrm{Off})>B_{\mathrm{byte}}$
    \textbf{ then } \Return $\textsc{Exec}(c,\mathcal{C})$
    \label{line:pace-budget}
    \Comment{Candidate set is too large}

\State $X\gets\textsc{Materialize}(D,\mathrm{Off},\mathcal{C})$
    \label{line:pace-materialize}
    \Comment{Preserve original corpus order}

\State \Return $\textsc{Exec}(c,X)$
    \label{line:pace-exec}
    \Comment{Apply exact command semantics to candidates}

\end{algorithmic}
\end{algorithm}

\subsection{Sharded-Parallel Corpus Search}
\label{app:eff-search-sharded}

\begin{algorithm*}[t]
\caption{Sharded-Parallel Corpus Search}
\label{alg:sharded-search}
\begin{algorithmic}[1]
\Require command containing pipe (|) $c$; corpus $\mathcal{C}$ split into $S$ contiguous shards $C_i$ s.t. $\biguplus_i C_i = \mathcal{C}$
\Ensure Byte-exact output identical to sequential execution of $c$ on $\mathcal{C}$

\State $(s_1,\dots,s_m) \gets \textsc{Decompose}(c)$ \Comment{Split pipeline into $m$ distinct stages based on pipe operator(|)}
\State $(\tau, N) \gets \textsc{Classify}(s_1,\dots,s_m)$ \Comment{Classify the type of reduction}
\State \textbf{if}{$\tau = \textsc{Sequential}$} \textbf{then} \Return $\textsc{Exec}(c, \mathcal{C})$ \Comment{Run on full corpus if it can only be sequentially}

\State \textbf{parallel for} $i=1,\ldots,S$ \textbf{do} $R_i \gets \textsc{Exec}(c, C_i)$ \Comment{Perform the operation on each shard in parallel}

\State \textbf{if} $\tau = \textsc{Concat}$ \textbf{then} \Return $\biguplus_i R_i$ \Comment{Concat result of operations}

\State \textbf{else if} $\tau = \textsc{Head}$ \textbf{then} \Return $\mathrm{TOP}(\biguplus_i R_i,\; N)$ \Comment{Concat result of operations and select top $N$}

\State \textbf{else if} $\tau = \textsc{COUNT}$ \textbf{then} \Return $\sum_i \mathrm{Int}(R_i)$ \Comment{Add results of operations}

\State \textbf{else if} $\tau = \textsc{SortHead}$ \textbf{then} \Return $\mathrm{TOP}(\mathrm{Merge}(R_1,\dots,R_S),\; N)$ \Comment{Merge results based on value and return top $N$}

\Statex
\Function{Classify}{$s_1,\dots,s_m$} \Comment{$m$ is the final stage of the pipeline}
    \If{$s_1 \notin \{\texttt{rg}, \texttt{grep}\} \lor \exists s_j \;\text{s.t.}\; \textsc{Unsafe}(s_j)$}
        \State \Return $(\textsc{Sequential}, \emptyset)$ \Comment{Reject non-search or cross-line context}
    \EndIf
    \State $\mathcal{S} \gets \{ s_i \in (s_1,\dots,s_m) \mid \textsc{Stateless}(s_i) \}$ \Comment{e.g., \texttt{cut}, \texttt{tr}, line-wise \texttt{sed}}
    \If{$s_m = \texttt{head -n } N \land \forall j < m,\; s_j \in \mathcal{S}$}
        \State \Return $(\textsc{Head}, N)$ \Comment{Pattern: Stateless maps $\to$ early termination}
    \ElsIf{$s_m = \texttt{wc -l} \land \forall j < m,\; s_j \in \mathcal{S}$}
        \State \Return $(\textsc{Count}, \emptyset)$ \Comment{Pattern: Stateless maps $\to$ global line count}
    \ElsIf{$s_{m-1} \in \{\texttt{sort},\; \texttt{sort|uniq}\} \land s_m = \texttt{head -n } N$}
        \State \Return $(\textsc{SortHead}, N)$ \Comment{Pattern: Stateless maps $\to$ Top-$K$ filter}
    \ElsIf{$\forall j \le m,\; s_j \in \mathcal{S}$}
        \State \Return $(\textsc{Concat}, \emptyset)$ \Comment{Pattern: Entire pipeline is purely stateless}
    \Else
        \State \Return $(\textsc{Sequential}, \emptyset)$ \Comment{Revert unsupported complex pipelines}
    \EndIf
\EndFunction
\end{algorithmic}
\end{algorithm*}


This appendix provides a comprehensive technical overview of the system-level optimizations employed by the DCI agent’s command-execution engine. The engine is designed under a strict correctness-first principle: any pipeline whose parallel execution cannot be guaranteed to be byte-identical to sequential execution is safely executed via a single-file fallback mechanism.

\subsubsection{Corpus Sharding and Parallel Fan-Out}

The primary bottleneck in evaluating shell pipelines over large-scale corpora (e.g., a 14\,GB JSONL file) is the inherently sequential execution model of standard Unix search tools. To address this, the engine performs a one-time, idempotent, line-aligned sharding of the corpus into $S$ disjoint partitions.

\begin{itemize}[leftmargin=1.2em]
    \item \textbf{Line-Aligned Sharding:} The corpus is partitioned along line boundaries (e.g., via \texttt{split -d -n l/S}), ensuring that each JSON record remains intact within a single shard. By construction, concatenating all shards in order reconstructs the original corpus without byte-level modification.

    \item \textbf{Thread-Level Fan-Out:} At inference time, shell pipelines are executed concurrently across the $S$ shards using a thread pool. Since each shard is processed via independent \texttt{subprocess.run} calls (which in turn invoke tools such as \texttt{rg}), threading avoids Python-level process management overhead while allowing all shard executions to proceed in parallel. This reduces end-to-end latency for full-corpus scans approximately proportional to the number of shards, up to the I/O and memory bandwidth limits of the system that is hosting the optimized engine.
\end{itemize}

\subsubsection{Pipeline Classification and Merge Strategies}

To guarantee exact behavioral equivalence with sequential corpus execution, the engine employs a conservative pipeline parser that dynamically classifies each shell pipeline and routes it to one of five deterministic execution strategies. If a pipeline begins with an unsupported primitive or contains stateful operations that violate shard independence---such as line-indexing flags (e.g., \texttt{-n}), count-based modes (e.g., \texttt{-c}), contextual windowing (e.g., \texttt{-A}, \texttt{-B}, \texttt{-C}), or in-place transformations (e.g., \texttt{sed -i})---it is immediately executed via the single-file fallback path. For supported pipelines, the engine aggregates the per-shard partial outputs $\{R_1, \dots, R_N\}$ using the following semantics:

\begin{description}
    \item[\texttt{CONCAT}:] Applied to fully stateless pipelines (e.g., \texttt{rg}, \texttt{grep}, \texttt{cut}, \texttt{tr}, \texttt{sed}). The engine executes the pipeline on each shard independently and concatenates the outputs $R_i$ in shard order.
    
    \item[\texttt{HEAD}:] Applied to pipelines terminating in \texttt{head -n K}. The engine applies the truncation locally per shard to bound memory usage to $K \times N$ lines. The bounded outputs are concatenated in shard order, and a final global \texttt{head -n K} is applied.
    
    \item[\texttt{COUNT}:] Applied to pipelines terminating in counting operations (e.g., \texttt{wc -l}). The engine extracts the scalar count from each shard and computes the global sum.
    
    \item[\texttt{SORTHEAD}:] Applied to top-$K$ retrieval pipelines containing \texttt{sort}, optionally \texttt{uniq}, and terminating in \texttt{head -n K}. The engine applies \texttt{sort | head -n K} to each shard. The resulting sorted streams are merged using a deterministic $k$-way merge (\texttt{sort -m}), followed by an optional global \texttt{uniq} and a final global \texttt{head -n K}.
    
    \item[\texttt{SEQUENTIAL}:] Applied to any unrecognized, unparseable, or globally stateful pipeline. The pipeline is executed sequentially against the unified single-file corpus to guarantee correctness.
\end{description}

\subsubsection{I/O and System-Level Optimizations}

\begin{table}[!t]
\centering
\caption{Hyperparameters used for synthetic cold-start trajectory generation (Algorithm~\ref{alg:coldstart}) and Supervised Fine-Tuning (SFT) of the DCI agent inside the \texttt{verl} FSDP training framework.}
\label{tab:hyperparams-sft}
\begin{tabular}{llc}
\toprule
\textbf{Phase} & \textbf{Hyperparameter} & \textbf{Value} \\
\midrule
\textbf{Cold-Start} 
& Tutor ($\mathcal{M}_T$) \& Planner ($\mathcal{M}_P$) Backbone & Qwen3.5-27B \\
\textbf{Data} 
& Max Refinement Iterations ($M$) & 5 \\
\textbf{Generation} 
& SFT Dataset Size & 10,000 \\
& Default Top-$p$ & 1.0 \\
& Tutor (Backward Phase) Temperature & $0.4 + (0.1 \times \text{iteration})$ \\
& Planner (Forward Phase) Temperature & $0.7$ \\
& Judge Phase Temperature & $0.6$ \\
\midrule
\textbf{Supervised} 
& Policy Model ($\pi_\theta$) & Qwen3.5-9B \\
\textbf{Fine-Tuning} 
& Epochs & 1 \\
\textbf{(SFT)} 
& Optimizer & AdamW \\
& Optimizer Betas ($\beta_1, \beta_2$) & (0.9, 0.999) \\
& Optimizer Epsilon ($\epsilon$) & $1 \times 10^{-8}$ \\
& Peak Learning Rate & $5 \times 10^{-6}$ \\
& Learning Rate Scheduler & Constant with Warmup \\
& Linear Warmup Ratio & 0.05 \\
& Weight Decay & 0.01 \\
& Gradient Clipping Norm & 1.0 \\
& Max Sequence Length & 16,384 \\
& Global Batch Size & 32 \\
& Precision & \texttt{bfloat16} \\
& Hardware Parallelism & Ulysses (size = 4) \\
\bottomrule
\end{tabular}
\end{table}

Because the engine predominantly evaluates fixed-string filtering operations, retrieval latency is largely determined by memory bandwidth and I/O access patterns. To maximize throughput, the system implements a tiered I/O optimization stack.

\paragraph{RAM-Resident Corpus Placement:} When system memory permits, the corpus and all shards are staged in a RAM-backed filesystem (e.g., \texttt{/dev/shm}). This ensures that all reads are served directly from main memory, avoiding filesystem and disk latency. In addition, the engine proactively warms the page cache at startup to eliminate cold-start penalties on the first retrieval query.

\paragraph{Deterministic Execution Flags:} The engine injects a set of deterministic performance flags into supported Unix tools. Memory-mapped I/O (e.g., \texttt{--mmap} for \texttt{rg} and \texttt{grep}) reduces system-call overhead, while \texttt{--no-config} disables user-level configuration to ensure reproducibility. The environment is also fixed to \texttt{LC\_ALL=C}, enabling bytewise matching and avoiding locale-dependent overhead without affecting literal search semantics.

\subsubsection{Persistent Daemon Architecture and Telemetry}

To eliminate the recurring costs of Python wrapper initialization and process startup across multiple tool calls within a single trajectory, the execution engine is deployed as a long-running persistent daemon. The evaluation loop communicates with the persistent \texttt{ShardedSearchEngine} via a length-prefixed JSON protocol over a Unix socket, amortizing per-call overhead and reducing latency by approximately 1--3 milliseconds per invocation. Finally, the engine preserves standard Unix failure semantics (e.g., returning exit code 0 if any shard produces a match) and deduplicates standard error streams to avoid redundant global error logging. Each tool invocation additionally records fine-grained telemetry---including the parsed command, selected merge strategy, shard configuration, and fallback decision---enabling detailed analysis of fast-path utilization during inference.


\subsection{Reward Function}
\label{app:reward}

\paragraph{Correctness reward:}

Let $\hat{y}$ denote the agent's final answer, extracted from the last
$\texttt{<answer>} \cdots \texttt{</answer>}$ block in the trajectory, and let
$\mathcal{Y}=\{y_1,\dots,y_m\}$ be the set of gold reference answers. We compute
a token-level $F_1$ score between $\hat{y}$ and the reference set. Following
prior work \citep{rajpurkar-etal-2016-squad}, each prediction and reference
answer is normalized by lowercasing, removing punctuation, dropping articles
(``a'', ``an'', ``the''), and tokenizing on whitespace. For a normalized prediction $\hat{y}$ and a reference answer $y \in \mathcal{Y}$,
let $P$ and $G$ denote their respective token multisets. We define the overlap as
the multiset intersection
$o = \sum_t \min(P(t), G(t))$, where $P(t)$ and $G(t)$ denote token counts. If $|P|=0$ or $|G|=0$, we set $F_1(\hat{y}, y)=0$. Otherwise, precision, recall,
and $F_1$ are defined as:
\begin{equation}
\mathrm{p} = \frac{o}{|P|}, \quad
\mathrm{r} = \frac{o}{|G|}, \quad
F_1(\hat{y}, y) = \frac{2\,\mathrm{p}\cdot\mathrm{r}}{\mathrm{p}+\mathrm{r}}.
\end{equation}
Since multiple surface forms may be valid, the final answer reward
$R_{\mathrm{ans}} \in [0,1]$ is defined as the maximum score over all reference
answers:
\begin{equation}
R_{\mathrm{ans}} = \max_{y \in \mathcal{Y}} F_1(\hat{y}, y).
\end{equation}
This provides a dense learning signal that assigns partial credit to partially
correct answers, rather than relying on a sparse binary exact-match reward.

\begin{table}[!t]
\centering
\caption{Hyperparameters used for reinforcement learning optimization via Group Relative Policy Optimization (GRPO) inside the \texttt{verl} training framework.}
\label{tab:hyperparams-grpo}
\begin{tabular}{llc}
\toprule
\textbf{Phase} & \textbf{Hyperparameter} & \textbf{Value} \\
\midrule
\textbf{GRPO} 
& Group Size ($n$) & 5 \\
\textbf{Algorithm} 
& PPO Clip Ratio ($\epsilon$) & 0.2 \\
& KL Divergence Coefficient & 0.0 (Disabled) \\
& PPO Epochs & 1 \\
& Policy Entropy Coefficient & 0.0 \\
\midrule
\textbf{Rollout \&} 
& Sampling Temperature & 1.0 \\
\textbf{Sampling} 
& Sampling Top-$p$ & 1.0 \\
& Max Sequence Length & 16,384 \\
& Max Assistant Turns & 6 \\
\midrule
\textbf{Optimization} 
& Optimizer & AdamW \\
\textbf{\& Batching} 
& Optimizer Betas ($\beta_1, \beta_2$) & (0.9, 0.999) \\
& Optimizer Epsilon ($\epsilon$) & $1 \times 10^{-8}$ \\
& Peak Learning Rate & $5 \times 10^{-6}$ \\
& Learning Rate Scheduler & Constant with Warmup \\
& Linear Warmup Ratio & 0.05 \\
& Weight Decay & 0.0 \\
& Gradient Clipping Norm & 1.0 \\
& Global Training Batch Size & 256 questions \\
& PPO Mini-batch Size & 32 \\
& PPO Micro-batch Size per GPU & 1 \\
& Precision & \texttt{bfloat16} \\
& Hardware Parallelism & Ulysses (size = 2) \\
& Total Training Steps & 200 \\
\bottomrule
\end{tabular}
\end{table}

\paragraph{Format reward:}

We define a binary format indicator $\phi \in \{0,1\}$ to evaluate the structural validity of a trajectory. A rollout is valid ($\phi=1$) if and only if it satisfies three strict formatting criteria:
(1) all special tags (e.g., \texttt{<think>}, \texttt{<tool\_call>}, \texttt{<tool\_response>}, and \texttt{<answer>}) are properly balanced and non-overlapping;
(2) the trajectory follows the prescribed state-transition structure (reasoning $\rightarrow$ tool invocation $\rightarrow$ environment response $\rightarrow$ final answer), with no text generated outside the designated blocks; and
(3) the trajectory terminates within a closing \texttt{</answer>} tag.

This constraint mirrors the strictly formatted interaction protocol enforced during cold-start supervised fine-tuning (SFT), where all trajectories are structurally valid by construction. Consequently, $\phi$ acts as a gate that isolates and penalizes formatting violations introduced by the policy during reinforcement learning exploration phase.

\paragraph{Combined reward:}

The final reward during reinforcement learning optimization is defined as:
\begin{equation}
R = \phi \cdot R_{\mathrm{ans}}
  = \phi \cdot \max_{y \in \mathcal{Y}} F_1(\hat{y}, y).
\end{equation}
Thus, only structurally valid trajectories can receive a positive reward; malformed or out-of-format trajectories receive zero reward regardless of answer correctness.

\subsection{Experimental Settings \& Hyperparameters}
\label{app:experimental-setting}

\paragraph{SFT Training Phase:} 

Table~\ref{tab:hyperparams-sft} summarizes the configuration for both synthetic cold-start data generation and the subsequent Supervised Fine-Tuning (SFT) stage. For data generation, we use the 27B variant of Qwen3.5 as both the answer-aware Tutor ($\mathcal{M}_T$) and answer-blind Planner ($\mathcal{M}_P$) to construct a 10,000-trajectory dataset. During the Tutor’s backward discovery phase, we apply a dynamic temperature schedule $0.4 + 0.1 \times \text{iter}$ (up to $M=5$ refinement steps) with Nucleus Sampling \citep{Holtzman2020The} to encourage broader exploration when initial retrieval attempts fail. In contrast, the Planner’s forward assembly and the final coherence judge use fixed temperatures of $0.7$ and $0.6$, respectively, with top-$p=1.0$ across all stages.

For SFT, the 9B Qwen3.5 policy model ($\pi_\theta$) is trained for one epoch using AdamW \citep{loshchilov2018decoupled} with a peak learning rate of $5 \times 10^{-6}$ and a global batch size of 32. We use a linear warmup over the first 5\% of training steps followed by a constant learning rate schedule. The maximum sequence length is set to 16,384 tokens to accommodate long interaction trajectories, including tool calls and retrieved context. Training is performed in the \texttt{verl}\footnote{Available at: \url{https://github.com/verl-project/verl}} framework using FSDP \citep{zhao2023pytorchfsdpexperiencesscaling} with \texttt{bfloat16} precision and Ulysses sequence parallelism (degree 4). The experiments are conducted on 4 NVIDIA A100 (80GB) GPUs on a machine with 1024GB memory.

\begin{table}[!t]
\centering
\caption{Hyperparameters and configurations used during inference for FlashRAG datasets and BrowseComp-Plus.}
\label{tab:hyperparams-inference}
\begin{tabular}{lcc}
\toprule
\textbf{Hyperparameter} & \textbf{FlashRAG} & \textbf{BrowseComp-Plus} \\
\midrule
Maximum Assistant Turns & 6 & 16 \\
Decoding Temperature & 0.6 & 0.6 \\
Decoding Top-$p$ & 1.0 & 1.0 \\
\texttt{tool\_max\_tokens} & 2,048 & 6,000 \\
\midrule
Maximum Model Sequence Length & 16,384 & 131,072 \\
GPU Memory Utilization & 0.90 & 0.90 \\
Tensor Parallel Size & 2 & 2 \\
Hardware Infrastructure & $2\times$ NVIDIA A100 & $2\times$ NVIDIA A100 \\
\bottomrule
\end{tabular}
\end{table}

\paragraph{GRPO Training Phase:}

Table~\ref{tab:hyperparams-grpo} lists the configuration parameters employed during the reinforcement learning phase using Group Relative Policy Optimization (GRPO) inside the \texttt{verl} framework. The agent policy ($\pi_\theta$) is initialized from the checkpoint optimized during the Supervised Fine-Tuning (SFT) stage and is further trained for a total of 200 global steps. For each input query, the policy samples a group of $n=5$ independent trajectories to compute relative advantages. Trajectory rewards are determined by the continuous token-level $F_1$ score, multiplied by a binary formatting gate to strictly penalize structural violations. Computed rewards are centered around the group mean and normalized by their standard deviation, providing a comparative learning signal that optimizes the policy without requiring an explicit critic network. We use a symmetric PPO clip ratio of 0.2 and disable the explicit KL divergence penalty to the reference policy, running exactly 1 proximal epoch per training step to minimize policy degradation.

\begin{figure}
    \centering
    \includegraphics[width=0.9\linewidth]{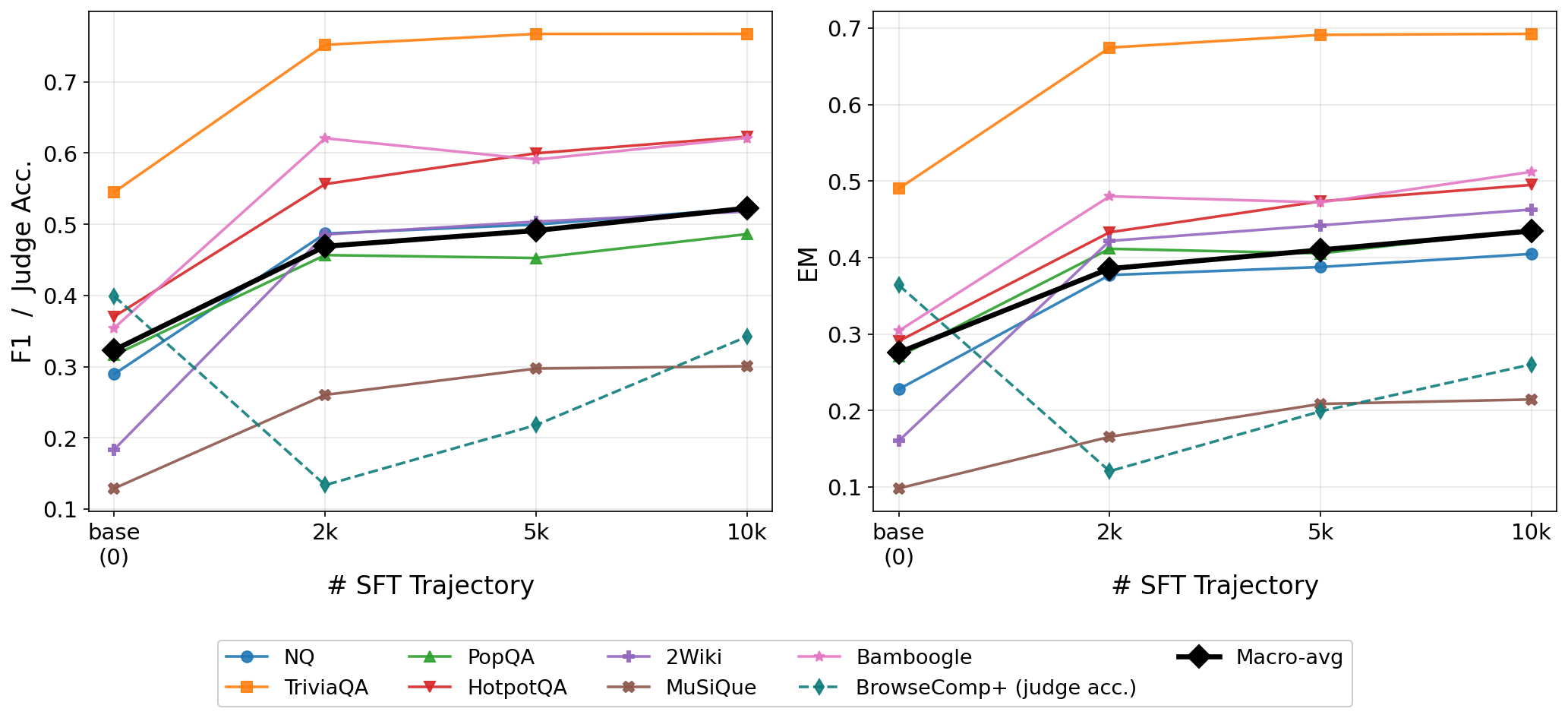}
    \vspace{-0.3cm}
    \caption{Effect of supervised fine-tuning (SFT) data size after RL training. The left panel shows FlashRAG $F_1$ and BrowseComp-Plus judged accuracy; the right panel shows EM for all eight datasets. Black lines show unweighted macro averages.}
    \label{fig:sft_em}
\end{figure}

During rollout generation via the vLLM engine,\footnote{Available at: \url{https://vllm.ai/}} sampling parameters are configured with a temperature of 1.0 and a top-$p$ of 1.0 to encourage diverse tool-use exploration while preserving syntax consistency. To support deep, multi-turn interactions over the corpus, the trajectory budget is capped at a maximum sequence length of 16,384 tokens and a maximum of 6 assistant turns. Policy updates are calculated using the AdamW optimizer with a peak learning rate of $5 \times 10^{-6}$ and a constant schedule following a 5\% linear warmup. The optimization runs across a global batch size of 256 questions per step, evaluated in mini-batches of 32 and micro-batches of 1 per GPU. Distributed processing is managed in \texttt{bfloat16} mixed precision utilizing a Ulysses sequence parallel size of 2 for both the actor and reference models. The experiments are conducted on a machine with 4 NVIDIA A100 (80GB) GPUs and 1024\,GB of system memory, taking approximately 4 days to complete.

\paragraph{Inference Setting:} 

Table~\ref{tab:hyperparams-inference} outlines the core configuration parameters and architectural settings used during the inference phase of our finalized DCI agent. At evaluation time, the agent policy leverages a decoding temperature restricted to $0.6$ alongside a top-$p$ of $1.0$ to ensure highly stable, syntactic consistency during multi-turn tool interaction. To prevent context space saturation from excessively broad document retrieval, the maximum length for any individual shell tool output is explicitly bounded via the \texttt{tool\_max\_tokens} parameter to $2,048$ tokens. In strict alignment with the environment constraints used during reinforcement learning, the maximum depth of a single interaction trajectory is capped at $6$ assistant turns, providing a consistent structural framework for the model's sequential reasoning. To maximize throughput and safely accommodate long-context interaction histories, the inference infrastructure relies on a maximum sequence length of $16,384$ tokens. The model is served using a specialized high-concurrency configuration capable of handling a maximum of $256$ sequences simultaneously, utilizing a high GPU memory allocation threshold of $0.90$. The physical compute layer is provisioned across a hardware infrastructure consisting of $2\times$ NVIDIA A100 GPUs, orchestrated with a tensor parallel size of $2$ to optimize distributed memory access and minimize generation latency during dense search rollouts.

\begin{table*}[!t]
    \centering
    \caption{Exact-match (EM) performance across datasets (Qwen3.5 backbone). Superscript $^*$ shows the datasets included in the training set, while all others are evaluated out-of-distribution. Superscript $^\uparrow$ indicates a statistically significant improvement, while $^\downarrow$ denotes a statistically significant decrease compared to the best-performing baseline. We use McNemar's test because significance is computed over paired binary exact-match outcomes for the same set of questions ($p < 0.05$).}
    \label{tab:em-result}
    \resizebox{\textwidth}{!}{
        \begin{tabular}{l l c c c c c c c c c}
        \toprule
        \multirow{2}{*}{\textbf{Method}} & \multirow{2}{*}{\textbf{Retriever}} & \multicolumn{3}{c}{\textbf{Single-hop}} & \multicolumn{5}{c}{\textbf{Multi-hop}} & \multirow{1}{*}{\textbf{Average}} \\
        \cmidrule(lr){3-5} \cmidrule(lr){6-10}
        & & \textbf{NQ$^*$} & \textbf{TriviaQA} & \textbf{PopQA} & \textbf{HotpotQA$^*$} & \textbf{2Wiki} & \textbf{MuSiQue} & \textbf{Bamboogle} & \textbf{\shortstack{BrowseComp+}} & (macro)\\

        \midrule

        Direct & --- & 0.1864 & 0.4917 & 0.1925 & 0.2063 & 0.2840 & 0.0401 & 0.1040 & 0.0012 & 0.1883 \\

        \midrule

        \multirow{3}{*}{RAG}
        & BM25 & 0.2471 & 0.5899 & 0.2762 & 0.3376 & 0.2923 & 0.0629 & 0.1760 & 0.0807 & 0.2578 \\
        & E5-110M & 0.4014 & 0.6264 & 0.3731 & 0.3213 & 0.2623 & 0.0753 & 0.2160 & 0.0361 & 0.2890 \\
        & Qwen3-4B & 0.3856 & 0.6365 & 0.4219 & 0.3492 & 0.2893 & 0.0840 & 0.2240 & 0.0904 & 0.3101 \\

        \midrule

        \multirow{3}{*}{IRCoT}
        & BM25 & 0.1828 & 0.4787 & 0.2322 & 0.2300 & 0.0992 & 0.0397 & 0.1040 & 0.0880 & 0.1818 \\
        & E5-110M & 0.3366 & 0.5416 & 0.3222 & 0.2248 & 0.0945 & 0.0592 & 0.1760 & 0.0241 & 0.2224 \\
        & Qwen3-4B & 0.3183 & 0.5655 & 0.3683 & 0.2591 & 0.1246 & 0.0707 & 0.2240 & 0.0831 & 0.2517 \\

        \midrule

        \multirow{3}{*}{Search-O1}
        & BM25 & 0.3125 & 0.6519 & 0.3449 & 0.3814 & 0.3033 & 0.1535 & 0.4160 & 0.1554 & 0.3398 \\
        & E5-110M & 0.3557 & 0.6527 & 0.3719 & 0.3639 & 0.2820 & 0.1456 & 0.4960 & 0.0410 & 0.3386 \\
        & Qwen3-4B & 0.3468 & 0.6524 & 0.4070 & 0.3761 & 0.3280 & 0.1688 & 0.4880 & 0.0699 & 0.3546 \\

        \midrule

        \multirow{3}{*}{\shortstack[l]{Rejection\\Sampling}}
        & BM25 & 0.2765 & 0.6415 & 0.3082 & 0.4374 & 0.3434 & 0.1680 & 0.4960 & 0.2313 & 0.3628 \\
        & E5-110M & 0.3111 & 0.6458 & 0.3488 & 0.4238 & 0.3142 & 0.1643 & 0.5360 & 0.1241 & 0.3585 \\
        & Qwen3-4B & 0.3183 & 0.6504 & 0.3776 & 0.4308 & 0.3668 & 0.1911 & 0.5440 & 0.1518 & 0.3789 \\

        \midrule

        \multirow{3}{*}{Search-R1}
        & BM25 & 0.3385 & 0.6993 & 0.3660 & 0.4512 & 0.3515 & 0.1734 & 0.4880 & 0.2446 & 0.3890 \\
        & E5-110M & 0.3823 & \textbf{0.7005} & 0.4118 & 0.4313 & 0.3202 & 0.1949 & 0.5840 & 0.1313 & 0.3945 \\
        & Qwen3-4B & 0.3884 & 0.6988 & \textbf{0.4503} & 0.4475 & 0.3823 & 0.2031 & \textbf{0.5920} & 0.1711 & 0.4167 \\

        \midrule
        \ourmethod & --- & \textbf{0.4047$^\uparrow$} & 0.6925 & 0.4363$^\downarrow$ & \textbf{0.4949$^\uparrow$} & \textbf{0.4627$^\uparrow$} & \textbf{0.2143} & 0.5120 & \textbf{0.2602} & \textbf{0.4347$^\uparrow$} \\
        \bottomrule
        \end{tabular}
    }
\end{table*}

\begin{table*}[!t]
    \centering
    \caption{Ablation study of \ourmethod{} (Qwen3.5 backbone) across single-hop and multi-hop datasets. We report F$_1$ for the FlashRAG datasets and judge accuracy for BrowseComp-Plus; EM results are in Table~\ref{tab:ablation_em} in Appendix~\ref{app:more-results}. Superscript $^\uparrow$ indicates a statistically significant improvement using Student's $t$-test over both ablated variants after Bonferroni correction.}
    \label{tab:ablation_f1}
    \resizebox{\textwidth}{!}{
        \begin{tabular}{l c c c c c c c c c}
        \toprule
        \multirow{2}{*}{\textbf{Variant}} 
        & \multicolumn{3}{c}{\textbf{Single-hop}} 
        & \multicolumn{5}{c}{\textbf{Multi-hop}} 
        & \multirow{2}{*}{\textbf{Average}} \\
        \cmidrule(lr){2-4} \cmidrule(lr){5-9}
        & \textbf{NQ} 
        & \textbf{TriviaQA} 
        & \textbf{PopQA} 
        & \textbf{HotpotQA} 
        & \textbf{2Wiki} 
        & \textbf{MuSiQue} 
        & \textbf{Bamboogle} 
        & \textbf{BrowseComp+} 
        & (macro) \\
        
        \midrule
        
        \ourmethod{} 
        & \textbf{0.5223$^\uparrow$} 
        & \textbf{0.7673$^\uparrow$} 
        & \textbf{0.4861$^\uparrow$} 
        & \textbf{0.6231$^\uparrow$} 
        & \textbf{0.5178$^\uparrow$} 
        & \textbf{0.3006$^\uparrow$} 
        & \textbf{0.6212$^\uparrow$} 
        & 0.3422
        & \textbf{0.5226} \\
        
        \midrule
        
        \quad - w/o GRPO 
        & 0.3880 
        & 0.6390 
        & 0.3903 
        & 0.4738 
        & 0.2956 
        & 0.2069 
        & 0.4231 
        & 0.2253
        & 0.3803 \\
        
        \quad - w/o SFT 
        & 0.2896 
        & 0.5451 
        & 0.3163 
        & 0.3705 
        & 0.1838 
        & 0.1291 
        & 0.3544 
        & \textbf{0.3988}
        & 0.3235 \\
        
        \bottomrule
        \end{tabular}
    }
\end{table*}

\begin{table*}[!b]
    \centering
    \caption{Ablation study of \ourmethod{} (Qwen3.5 backbone) across single-hop and multi-hop datasets (EM scores). Superscript $^\uparrow$ indicates a statistically significant improvement over both ablated variants. We use McNemar's test because significance is computed over paired binary exact-match outcomes for the same set of questions, and apply Bonferroni correction ($p < 0.05$).}
    \label{tab:ablation_em}
    \resizebox{\textwidth}{!}{
        \begin{tabular}{l c c c c c c c c c}
        \toprule
        \multirow{2}{*}{\textbf{Variant}} 
        & \multicolumn{3}{c}{\textbf{Single-hop}} 
        & \multicolumn{5}{c}{\textbf{Multi-hop}} 
        & \multirow{2}{*}{\textbf{Average}} \\
        \cmidrule(lr){2-4} \cmidrule(lr){5-9}
        & \textbf{NQ} 
        & \textbf{TriviaQA} 
        & \textbf{PopQA} 
        & \textbf{HotpotQA} 
        & \textbf{2Wiki} 
        & \textbf{MuSiQue} 
        & \textbf{Bamboogle} 
        & \textbf{BrowseComp+} 
        & (macro) \\
        
        \midrule
        
        \ourmethod{} 
        & \textbf{0.4047$^\uparrow$} 
        & \textbf{0.6925$^\uparrow$} 
        & \textbf{0.4363$^\uparrow$} 
        & \textbf{0.4949$^\uparrow$} 
        & \textbf{0.4627$^\uparrow$} 
        & \textbf{0.2143$^\uparrow$} 
        & \textbf{0.5120$^\uparrow$} 
        & 0.2602
        & \textbf{0.4347} \\
        
        \midrule
        
        \quad - w/o GRPO 
        & 0.2853 
        & 0.5696 
        & 0.3370 
        & 0.3679 
        & 0.2454 
        & 0.1332 
        & 0.3520 
        & 0.2133
        & 0.3130 \\
        
        \quad - w/o SFT 
        & 0.2277 
        & 0.4901 
        & 0.2704 
        & 0.2906 
        & 0.1601 
        & 0.0981 
        & 0.3040 
        & \textbf{0.3639}
        & 0.2756 \\
        
        \bottomrule
        \end{tabular}
    }
\end{table*}

\begin{figure*}[!t]
\begin{promptbox}[Backward command -- system]
You are an expert at writing shell commands to retrieve specific Wikipedia passages from a large JSONL corpus.

{corpus_description}

You are working BACKWARD from a known answer to construct a tool command that, when executed, returns a passage that confirms the answer to a sub-question.

CRITICAL — do not violate these:

(1) ANSWER-LEAK RULE.
Your shell command must NOT use the expected_answer string itself (or a near-identical paraphrase of it) as a search term. The agent at inference time will not know the answer; the answer must EMERGE from the document the command retrieves, not be an input to the search.

You are however ENCOURAGED to use:
- Words and phrases from the sub-question.
- Synonyms, paraphrases, and reformulations of those words. Pure keyword search often misses; trying near-synonyms is a real agent skill (e.g., "head office" / "headquartered in" / "based in" / "located in", "founded" / "established" / "started", "directed by" / "directed", "wife" / "spouse" / "married").
- Related concepts that an agent could plausibly guess from the sub-question alone.

The hard line: don't paste the literal answer string into the command.

Example (sub_question = "Which hotel company is the Oberoi family part of?", expected_answer = "The Oberoi Group"):
  WRONG (literal answer):           rg -F 'Oberoi Group' corpus.jsonl | head -n 3
  WRONG (near-identical paraphrase): rg -F 'The Oberoi' corpus.jsonl | head -n 3
  RIGHT (sub-question term):        rg -F 'Oberoi family' corpus.jsonl | rg -i -F 'hotel' | head -n 3
  RIGHT (synonym variation):        rg -F 'Oberoi family' corpus.jsonl | rg -i -F 'hospitality' | head -n 3
  RIGHT (broader, then narrow):     rg -F 'Oberoi family' corpus.jsonl | head -n 5

If keyword search keeps coming up empty, switch to a different paraphrase of the sub-question's key concept rather than reaching for the answer.

(2) OUTPUT SIZE.
Keep output short — typically `| head -n 3`. You may go up to `| head -n 8` when you need to scan more chunks of the same article (e.g., the article's intro is in chunk 1 but a specific fact is in a later chunk). If you're hitting clearly-irrelevant articles, prefer refining the QUERY before enlarging the head cap.

(3) STRUCTURE.
A single shell pipeline. No redirection (>), no chaining (; && ||), no command substitution (\$(...), `...`). Allowed tools: rg, grep, find, sed, awk, head, tail, cat, ls, wc, sort, cut, uniq, tr.
\end{promptbox}
\vspace{-0.4cm}
\caption{System instructions for the backward retrieval task, emphasizing the anti-leak rule.}
\label{fig:prompt_backward_system}
\end{figure*}

\begin{figure*}[!t]
\begin{promptbox}[Backward command -- first attempt]
Sub-question to find a supporting passage for: {sub_question}

Expected answer (known to you for verification only — do NOT use any of these as search terms):
- primary: {expected_answer}
- alternative forms (also forbidden as search terms): {forbidden_forms}

Other passages already retrieved for later sub-questions in this chain (for context only):
{downstream_docs}

Reason briefly about which Wikipedia article would most directly answer this sub-question, and what distinguishing terms from the SUB-QUESTION (not the answer or any of its alternative forms) you'd use to find it. Then output your single shell command.

Output exactly:
<reasoning>
your reasoning (2-4 sentences); explicitly check that your command does NOT include the expected answer or any of its alternative forms as a search term
</reasoning>
<command>
your single-pipeline shell command
</command>
\end{promptbox}
\vspace{-0.4cm}
\caption{Tutor prompt for the initial attempt at backward evidence retrieval.}
\label{fig:prompt_backward_first}
\end{figure*}

\newpage

\section{Additional Results}
\label{app:more-results}

In this section, we present supplementary Exact Match (EM) results for all eight datasets. While the main text reports token-level $F_1$ for FlashRAG and judged accuracy for BrowseComp-Plus, EM provides a stricter assessment of surface-form agreement. Importantly, the main conclusions from the $F_1$ analysis remain consistent under EM, indicating that the observed gains are not driven by partial-match artifacts. Table~\ref{tab:em-result} reports these results across all evaluated methods. \ourmethod{} achieves the highest overall macro-average EM score of $0.4347$, compared with $0.4167$ for the strongest baseline.

At the dataset level, \ourmethod{} achieves the best EM scores on five of the eight benchmarks (NQ, HotpotQA, 2Wiki, MuSiQue, and BrowseComp-Plus), with statistically significant gains on NQ, HotpotQA, and 2Wiki ($^\uparrow$). These results further support the effectiveness of direct corpus interaction for precise multi-hop reasoning and lexical evidence discovery. Conversely, EM also reflects the limitations of lexical search: \ourmethod{} shows a statistically significant drop on PopQA ($^\downarrow$) and lower performance on TriviaQA and Bamboogle, where surface-form variation and semantic matching can favor dense retrieval. Overall, the EM results reinforce the main finding that \ourmethod{} provides a precise and competitive alternative to index-based retrieval.

Tables~\ref{tab:ablation_f1} and~\ref{tab:ablation_em} report the ablation results under $F_1$ and EM, respectively. The full \ourmethod{} model achieves the highest macro-average performance under both metrics: $0.5226$ versus $0.3803$ without GRPO and $0.3235$ without SFT for $F_1$, and $0.4347$ versus $0.3130$ and $0.2756$ for EM. Removing GRPO therefore substantially degrades overall performance, while removing the SFT initialization produces an even larger drop, supporting the importance of structured trajectory bootstrapping before RL. One exception is BrowseComp-Plus, where the RL-only variant performs better than the full model; nevertheless, its aggregate performance remains substantially lower across the complete benchmark suite. Figure~\ref{fig:ablation_scaling} further studies the effect of cold-start SFT data size using both $F_1$ and EM. Policies are initialized with $0$, $2$k, $5$k, or $10$k trajectories and then optimized using the same GRPO configuration. Both metrics exhibit the same overall trend: introducing only $2$k cold-start trajectories yields a large improvement over RL from the base model, while increasing the dataset to $5$k and $10$k provides smaller but consistent gains. The improvements begin to plateau beyond $5$k examples, suggesting that the primary benefit of cold-start supervision comes from establishing structured search behavior before RL rather than simply scaling the amount of supervised data.

\begin{table}[t]
\centering
\caption{Evolution of trajectory characteristics during RL training of \ourmethod{} (Qwen3.5 backbone).}
\small
\setlength{\tabcolsep}{2.5pt}
\begin{adjustbox}{max width=\linewidth}
\begin{tabular}{l c c c c c}
\toprule
\textbf{RL} & \textbf{Cmds /} & \textbf{Lines Scanned} & \textbf{Resp.} & \textbf{Filter} & \textbf{Pipes /}\\
\textbf{Step} & \textbf{Traj.} & (\texttt{head\,-n}) & \textbf{Tokens} & \textbf{Chain} & \textbf{Cmd}\\
\midrule
10  & 3.06 & 4.9  & 4251 & 79\% & 1.98\\
50  & 2.59 & 6.9  & 4554 & 80\% & 2.04\\
100 & 2.56 & 11.8 & 5979 & 79\% & 2.04\\
200 & 2.56 & 10.4 & 6409 & 78\% & 2.01\\
\bottomrule
\end{tabular}
\end{adjustbox}
\label{tab:cmd-evo}
\end{table}

Table~\ref{tab:f1-result-4b} evaluates \ourmethod{} with the smaller Qwen3-4B-thinking backbone against Search-R1 using BM25, E5, and Qwen3-Emb-4B retrievers. (Note that ``Qwen3-4B'' here denotes the shared policy backbone, distinct from the Qwen3-Emb-4B embedding retriever used by the Search-R1 baselines.) \ourmethod{} achieves the highest macro-average score of $0.4280$, compared with $0.3946$ for the strongest Search-R1 variant. The gains are concentrated on multi-hop tasks: \ourmethod{} achieves the best performance on HotpotQA, 2Wiki, MuSiQue, and BrowseComp-Plus, with a particularly large improvement on BrowseComp-Plus ($0.1542$ vs.\ $0.0578$ for the best baseline). In contrast, retrieval-based baselines remain stronger on several single-hop datasets, including NQ, TriviaQA, and PopQA, as well as Bamboogle. Overall, these results show that the advantages of DCI, particularly for multi-step evidence discovery, persist when using a substantially smaller backbone.

\begin{table*}[t]
\centering
\caption{Performance across datasets (Qwen3-4B-thinking policy backbone for both \ourmethod{} and Search-R1; ``Qwen3-Emb-4B'' in the Retriever column is a separate embedding retriever, not the backbone). We report $F_1$ for FlashRAG datasets and GPT-4.1-judged accuracy for BrowseComp-Plus. Superscript $^*$ denotes datasets used in training.}
\label{tab:f1-result-4b}

\resizebox{\textwidth}{!}{
\begin{tabular}{llccccccccc}
\toprule
\textbf{Method} &
\textbf{Retriever} &
\multicolumn{3}{c}{\textbf{Single-hop}} &
\multicolumn{5}{c}{\textbf{Multi-hop}} &
\textbf{Average} \\
\cmidrule(lr){3-5}
\cmidrule(lr){6-10}
& &
\textbf{NQ$^*$} &
\textbf{TriviaQA} &
\textbf{PopQA} &
\textbf{HotpotQA$^*$} &
\textbf{2Wiki} &
\textbf{MuSiQue} &
\textbf{Bamboogle} &
\textbf{\shortstack{BrowseComp+}} &
\textbf{(macro)} \\
\midrule

\multirow{3}{*}{Search-R1} & BM25 & 0.3681 & 0.6532 & 0.3398 & 0.3877 & 0.3732 & 0.1537 & 0.4194 & 0.0578 & 0.3441 \\
 & E5-110M & \textbf{0.4608} & \textbf{0.6907} & 0.3996 & 0.4032 & 0.3833 & 0.1704 & \textbf{0.5038} & 0.0253 & 0.3796 \\
 & Qwen3-Emb-4B & 0.4552 & 0.6867 & \textbf{0.4636} & 0.4099 & 0.4184 & 0.1820 & 0.4980 & 0.0434 & 0.3946 \\

\midrule

\ourmethod & --- & 0.4544 & 0.6831 & 0.4308 & \textbf{0.5562} & \textbf{0.4307} & \textbf{0.2340} & 0.4808 & \textbf{0.1542} & \textbf{0.4280} \\

\bottomrule
\end{tabular}
}

\vspace{-0.4cm}
\end{table*}

\newpage
\section{Case Studies}
\label{app:case-studies}

In this section, we present qualitative case studies comparing the reasoning and retrieval trajectories of our DCI agent (\ourmethod{}) against the strongest dense retrieval baseline, Search-R1 with the Qwen3-Emb-4B retriever. These examples illustrate the different behaviors, strengths, and failure modes of direct corpus interaction through shell commands versus embedding-based semantic retrieval. For each example, we provide the original question, the ground-truth answer, and the generated reasoning traces, retrieval commands, and retrieved observations produced by both systems. The examples highlight two key phenomena observed throughout evaluation:

\begin{itemize}[leftmargin=1.2em]
    \item \textbf{Lexical Precision and Multi-Hop Evidence Isolation:} \ourmethod{} performs particularly well in settings that require exact lexical matching and iterative evidence filtering. Using shell operators such as \texttt{rg} and \texttt{grep}, the agent can isolate rare symbolic strings, progressively refine intermediate retrieval results, and compose multi-stage retrieval pipelines that accurately bridge evidence across documents (see Examples \ref{ex:hotpot-formula}, \ref{ex:hotpot-rockstar}, \ref{ex:hotpot-faithlutheran}, and \ref{ex:hotpot-joggers}). This behavior is especially beneficial for multi-hop reasoning, entity disambiguation, and cases where small lexical details determine correctness. In contrast, dense retrievers may smooth over these distinctions due to embedding-level semantic compression, sometimes leading to incorrect generalization or entity confusion.
    
    \item \textbf{Surface-Form Sensitivity and Ranking Limitations:} At the same time, direct corpus interaction inherits the limitations of lexical retrieval. Since shell-based search lacks a learned semantic ranking mechanism, relevant documents may appear later in the retrieval stream despite containing the correct evidence (see Example \ref{ex:bamboogle-citibank}). Moreover, strict surface-form matching can make the agent sensitive to lexical variations such as spelling differences or omitted diacritics, causing failures in cases where dense retrievers naturally generalize through semantic similarity (see Example \ref{ex:popqa-vaillant}).
\end{itemize}

\paragraph{Discussion of Qualitative Examples:} Here we discuss and explain the provided case studies:

\begin{itemize}
    \item \textbf{Symbolic and Rare-Token Matching (Example \ref{ex:hotpot-formula}):} This example demonstrates \ourmethod{}'s ability to locate highly specific strings like chemical formulas that dense retrievers often struggle to represent in vector space. While the dense retriever returns semantically ``chemistry-adjacent'' documents, it fails to identify the exact formula. Our agent uses \texttt{rg -F} for an exact match, proving the utility of direct symbolic interaction for technical queries.
    
    \item \textbf{Entity Precision and Disambiguation (Examples \ref{ex:hotpot-rockstar} and \ref{ex:hotpot-faithlutheran}):} These cases illustrate how lexical precision prevents the ``semantic collapse'' often seen in dense models. In Example \ref{ex:hotpot-rockstar}, our agent distinguishes between a subsidiary studio and its parent brand. In Example \ref{ex:hotpot-faithlutheran}, the agent uses the full, unique school name to isolate a single location, whereas the dense retriever retrieves a different school with a similar name, causing the reasoning to cascade into a geographical error.
    
    \item \textbf{Multi-Hop Evidence Composition (Example \ref{ex:hotpot-joggers}):} Here, we highlight the agent's ability to filter retrieval results iteratively. By searching for a specific band and then filtering for ``singer,'' the agent isolates the relevant person. It then effectively parses a ``highest [noun]'' construction, showing that DCI can handle complex, multi-step queries by treating the corpus as a structured database.
    
    \item \textbf{Temporal and Stale Information (Example \ref{ex:bamboogle-skytree}):} This highlights the importance of precise record-keeping. Our agent retrieves the document explicitly referencing the ``current'' record holder, while the dense retriever is misled by a document containing stale information about the previous record holder (CN Tower).
    
    \item \textbf{Limitations of Ranking and Surface Forms (Examples \ref{ex:bamboogle-citibank} and \ref{ex:popqa-vaillant}):} These examples expose the failure modes of DCI. In Example \ref{ex:bamboogle-citibank}, the lack of ranking forces the agent to rely on file order, burying the correct answer. In Example \ref{ex:popqa-vaillant}, we see that if the agent does not guess the exact surface form (e.g., handling diacritics or name variants), it fails where semantic models would intuitively generalize, underscoring the trade-off between lexical control and semantic robustness. By contrast, Example \ref{ex:popqa-arndt} is a case where the exact-name match succeeds precisely where dense retrieval blurs two similarly named entities (see the entity-precision discussion above).
\end{itemize}

\begin{cmpbox}[label={ex:hotpot-formula},title={\winbadge\ \ \textsc{HotpotQA -- symbolic / rare-token match}}]
\textbf{Q:} Which component of shellite has the formula (ON)CHOH?\\[1pt]
\textbf{Gold:} picric acid\par\smallskip
\textbf{Ours (grep agent)}~\cmark\par
  \thk{I need to identify the specific component of shellite that corresponds to the formula (ON)CHOH. Since no search has been run yet, I will first query the corpus for "shellite" to locate its definition and list its constituent components.}
  \cmd{rg -F "shellite" corpus.jsonl | head -n 8}
  \retr{"Glossary of British ordnance terms" TNT. After World War I the term "common lyddite" was dropped, and remaining stocks of lyddite-filled shells were referred to as HE (high explosive) shell filled lyddite. Hence "common" faded from use, replaced by "HE" as the explosive shell designation. Com ...}
  \thk{The search results confirm that Shellite is an explosive mixture of picric acid and dinitrophenol (or hexanitrodiphenylamine). To determine which of these components corresponds to the formula (ON)CHOH, I need to search the corpus for this specific chemical notation.}
  \cmd{rg -F "(ON)CHOH" corpus.jsonl | head -n 8}
  \retr{"Picric acid" Picric acid Picric acid is an organic compound with the formula (ON)CHOH. Its IUPAC name is 2,4,6-trinitrophenol (TNP). The name "picric" comes from the Greek ("pikros"), meaning "bitter", reflecting its bitter taste. It is one of the most acidic phenols. Like other highly ...}
  \thk{The search results confirm that the chemical formula (ON)CHOH corresponds to picric acid. The earlier search established that Shellite is an explosive mixture primarily composed of picric acid and dinitrophenol. Therefore, the component of shellite with the formula (ON)CHOH is picric acid.}
  \ans{picric acid}
\smallskip\par\textbf{Search-R1 + Qwen3-Emb-4B}~\xmark\par
  \thk{The user is asking about the chemical composition of Shellite, specifically which component has the formula (ON)CHOH. Let me search for information about Shellite and its components.}
  \cmd{search 'Shellite component formula (ON)CHOH'}
  \retr{Doc 1 (Title: Limpet) matrix, this can have varying profound effects on the formation of the goethite crystals. The space in between the crystals and the chitin matrix is filled with an amorphous hydrated silica (SiO). The most prominent metal by percent composition is iron in the form of goethite. ...}
  \thk{The search results don't directly answer the question about Shellite and the formula (ON)CHOH. Let me search more specifically about Shellite and its components.}
  \cmd{search 'Shellite explosive components formula'}
  \retr{Doc 1 (Title: "Shellite (explosive)") Shellite (explosive) Shellite (known as Tridite in US service) is an explosive mixture of picric acid and dinitrophenol or picric acid and hexanitrodiphenylamine in a ratio of 70/30. It was typically used as a filling in Royal Navy armour-piercing shells during ...}
  \thk{The search results tell me that Shellite is an explosive mixture of picric acid and dinitrophenol or picric acid and hexanitrodiphenylamine in a ratio of 70/30. However, I need to find which component has the formula (ON)CHOH. Let me search more specifically about the formula (ON)CHOH and its relation to Shellite components.}
  \cmd{search '(ON)CHOH chemical formula compound'}
  \retr{Doc 1 (Title: CH3O) CH3O The molecular formula CHO may refer to: The chemical name for CH3O is methoxide. It is a base formed from methanol by replacement of the hydroxyl hydrogen with a metal. A strong base and a good nucleophile. Methoxide contains one atom of carbon, three atoms of hydrogen and o ...}
  \thk{The search results don't directly give me the answer about Shellite and the formula (ON)CHOH. Let me search more specifically about dinitrophenol or hexanitrodiphenylamine, which are components of Shellite, to see if either of them matches the formula (ON)CHOH. Actually, looking at the formula (ON)CHOH, this looks like it might be referring to a specific compound. Let me search more specifically about this formula in relation to explosives or Shellite components.}
  \cmd{search 'Shellite dinitrophenol formula (ON)CHOH'}
  \retr{Doc 1 (Title: 2,4-Dinitrophenol) 2,4-Dinitrophenol 2,4-Dinitrophenol (2,4-DNP or simply DNP) is an organic compound with the formula HOCH(NO). It is a yellow, crystalline solid that has a sweet, musty odor. It sublimes, is volatile with steam, and is soluble in most organic solvents as well as aqueo ...}
  \thk{Based on the search results, I can see that: 1. Shellite is a mixture of picric acid and dinitrophenol (or picric acid and hexanitrodiphenylamine) 2. 2,4-Dinitrophenol has the formula HOCH(NO), which can be written as (ON)CHOH The formula (ON)CHOH matches the formula for 2,4-Dinitrophenol (also known as DNP), which is one of the components of Shellite. So the answer should be dinitrophenol (or specifically 2,4-dinitrophenol).}
  \ans{dinitrophenol}
\smallskip\par\textbf{Why:} \textit{The answer is keyed by a literal symbolic string. Lexical grep matches it exactly; a dense retriever cannot embed a chemical formula and returns chemistry-shaped but wrong passages.}
\end{cmpbox}\bigskip

\begin{cmpbox}[label={ex:hotpot-rockstar},title={\winbadge\ \ \textsc{HotpotQA -- entity precision (subsidiary vs.\ parent)}}]
\textbf{Q:} Paul DeBoy is known for an appearance in the Western action-adventure game developed by which company?\\[1pt]
\textbf{Gold:} Rockstar San Diego\par\smallskip
\textbf{Ours (grep agent)}~\cmark\par
  \thk{To answer the question, I first need to identify which Western action-adventure game Paul DeBoy is known for appearing in. I will search the Wikipedia corpus for "Paul DeBoy" to find details about his filmography and any specific video game roles.}
  \cmd{rg -F "Paul DeBoy" corpus.jsonl | head -n 8}
  \retr{"PJ DeBoy" PJ DeBoy Paul J. "PJ" DeBoy (born June 7, 1971) is an American actor and talk show host. DeBoy was born in Baltimore, Maryland. He shares his name with his cousin, actor Paul DeBoy. He and Paul are cousins of Delegate Steven J. DeBoy Sr. DeBoy started his career in New York, where he pe ...}
  \thk{The search results confirm that Paul DeBoy is best known for his appearance in "Red Dead Redemption" as Jimmy Saint. The question describes this as a "Western action-adventure game," which matches the genre of *Red Dead Redemption*. To answer the question, I now need to identify the company that developed this specific game.}
  \cmd{rg -F "Red Dead Redemption" corpus.jsonl | rg -i -F "developed" | head -n 8}
  \retr{"Red Dead" Red Dead Red Dead is a series of Western-themed action-adventure video games published by Rockstar Games. The first entry in the series, "Red Dead Revolver", was released on the PlayStation 2 and Xbox on May 3, 2004. Originally developed by Capcom, "Red Dead Revolver" borrowed element ...}
  \thk{Based on the search results, Paul DeBoy is known for his appearance in the Western action-adventure game *Red Dead Redemption*, where he played the character Jimmy Saint. The corpus confirms that *Red Dead Redemption* was developed by Rockstar San Diego (a subsidiary of Rockstar Games).}
  \ans{Rockstar San Diego}
\smallskip\par\textbf{Search-R1 + Qwen3-Emb-4B}~\xmark\par
  \thk{The user is asking about Paul DeBoy and which company developed a Western action-adventure game that he appeared in. Let me search for information about Paul DeBoy and his video game appearances.}
  \cmd{search 'Paul DeBoy video game appearance Western action-adventure game'}
  \retr{Doc 1 (Title: "Paul DeBoy") Paul DeBoy Paul DeBoy (born September 14, 1955 in Baltimore, Maryland) is an American actor. He is best known for appearances in "A Dirty Shame" as Wendell Doggett, "Red Dead Redemption" as Jimmy Saint, "Haber" as Bernhard Moritz and for episodes of "Law \& Order" and "Law ...}
  \thk{Based on the search results, I can see that Paul DeBoy is known for appearing in "Red Dead Redemption" as Jimmy Saint. Red Dead Redemption is a Western action-adventure game. I need to find out which company developed Red Dead Redemption. Let me search for more information about Red Dead Redemption's developer.}
  \cmd{search 'Red Dead Redemption developer company'}
  \retr{Doc 1 (Title: "Red Dead Redemption 2") Sadie, and Charles. Afterward, Sadie and Charles leave for other pursuits. The final scene shows Pinkerton agents Ross and Fordham observing John's ranch, initiating the events of "Red Dead Redemption". "Red Dead Redemption 2" was developed by Rockstar Studios, ...}
  \thk{Based on my search results, I can see that: 1. Paul DeBoy appeared in "Red Dead Redemption" as Jimmy Saint (according to Doc 1) 2. Red Dead Redemption is a Western-themed action-adventure game 3. The game was developed by Rockstar San Diego and published by Rockstar Games (according to Doc 3) The question asks which company developed the game. The game was developed by Rockstar San Diego (a studio of Rockstar Games). The main company is Rockstar Games. So the answer should be Rockstar Games.}
  \ans{Rockstar Games}
\smallskip\par\textbf{Why:} \textit{Ours retrieves the precise developing studio; the dense baseline collapses to the better-known parent brand.}
\end{cmpbox}\bigskip

\begin{cmpbox}[label={ex:hotpot-faithlutheran},title={\winbadge\ \ \textsc{HotpotQA -- name-collision disambiguation}}]
\textbf{Q:} What major city is the Faith Lutheran Middle School and High School located by?\\[1pt]
\textbf{Gold:} Las Vegas\par\smallskip
\textbf{Ours (grep agent)}~\cmark\par
  \thk{I need to first identify the specific location of Faith Lutheran Middle School and High School before I can determine which major city it is near. I will search the corpus for the school's name to find its entry and extract the location details.}
  \cmd{rg -F "Faith Lutheran Middle School and High School" corpus.jsonl | head -n 8}
  \retr{"Vernon Fox" Vernon Fox Vernon Lee Fox, III (born October 9, 1979) is a retired American football safety. He was signed by the San Diego Chargers as an undrafted free agent after the 2002 NFL Draft. He played college football at Fresno State. Fox is currently a professional motivational speaker. In ...}
  \thk{The search results confirm that Faith Lutheran Middle School and High School is located in Las Vegas, Nevada. Las Vegas is a major city in the United States, known for being the county seat of Clark County and a global hub for tourism and entertainment.}
  \ans{Las Vegas}
\smallskip\par\textbf{Search-R1 + Qwen3-Emb-4B}~\xmark\par
  \thk{The user is asking about the location of Faith Lutheran Middle School and High School. I need to search for this information to find out which major city this school is located near. Let me search for information about Faith Lutheran Middle School and High School to find its location.}
  \cmd{search 'Faith Lutheran Middle School and High School location city'}
  \retr{Doc 1 (Title: "Faith Lutheran High School") Faith Lutheran High School Faith Lutheran High School, commonly referred to as Faith or FLHS, is a private high school located in Crystal Lake, Illinois. It is affiliated with the Lutheran Church–Missouri Synod and serves students of McHenry County, Illino ...}
  \thk{The result is Faith Lutheran High School in Crystal Lake, Illinois -- but the question names "Faith Lutheran Middle School and High School", and Crystal Lake is not a major city. The nearest major city would be Chicago. Let me search again in case a differently named school matches. […]}
  \cmd{search 'Faith Lutheran Middle School and High School location city Illinois'}
  \retr{Doc 1 (Title: "Faith Lutheran High School") Faith Lutheran High School Faith Lutheran High School, commonly referred to as Faith or FLHS, is a private high school located in Crystal Lake, Illinois. It is affiliated with the Lutheran Church–Missouri Synod and serves students of McHenry County, Illino ...}
  \thk{Same result again -- Crystal Lake, Illinois. Since Crystal Lake is not a major city, the nearest one is Chicago. Let me try once more in case the intended school is elsewhere. […]}
  \cmd{search 'Faith Lutheran Middle School and High School location Chicago'}
  \retr{Doc 1 (Title: "Faith Lutheran High School") Faith Lutheran High School Faith Lutheran High School, commonly referred to as Faith or FLHS, is a private high school located in Crystal Lake, Illinois. It is affiliated with the Lutheran Church–Missouri Synod and serves students of McHenry County, Illino ...}
  \thk{All results return Faith Lutheran High School in Crystal Lake, Illinois, which I take to be the same institution. Crystal Lake is not a major city, so the question is probably asking for the nearest one; it lies in the Chicago metropolitan area, so the answer would be Chicago. […]}
  \ans{Chicago}
\smallskip\par\textbf{Why:} \textit{The exact full name pins the right entity in one hop; dense retrieval grabs a same-named school and the error cascades into a wrong city.}
\end{cmpbox}\bigskip

\begin{cmpbox}[label={ex:hotpot-joggers},title={\winbadge\ \ \textsc{HotpotQA -- multi-hop bridge + distinctive phrase}}]
\textbf{Q:} The Joggers are a four-piece band whose lead singer is the son of an American chemist who received the highest what?\\[1pt]
\textbf{Gold:} Hirsch index rating\par\smallskip
\textbf{Ours (grep agent)}~\cmark\par
  \thk{I need to identify the lead singer of the band "The Joggers" to determine who the son of an American chemist is. I will search the corpus for "The Joggers" and filter for mentions of "singer" to quickly locate the relevant biographical details.}
  \cmd{rg -F "The Joggers" corpus.jsonl | rg -i -F "singer" | head -n 8}
  \retr{"George T. Whitesides" is married to Loretta Hidalgo Whitesides, and they plan to be the first couple to honeymoon in space. His father is George M. Whitesides, a professor of chemistry at Harvard University. His brother is the lead singer of The Joggers, a Portland, Oregon-based band. George T. Whi ...}
  \thk{The search results confirm that Ben Whitesides is the lead singer of The Joggers and the son of George M. Whitesides, an American chemist. The question asks for the "highest what" he received, which likely refers to a specific, top-tier honor or distinction in his field. Since the snippet mentions he "received the highest Hirsch index rating," this might be a distractor or a specific detail, but the question phrasing "highest [noun]" suggests a standard award title like "highest honor" or "highest award." To find the specific answer, I should search for George M. Whitesides in the context of awards or honors, filtering for the word "highest" to match the question's phrasing.}
  \cmd{rg -F "George M. Whitesides" corpus.jsonl | rg -i -F "highest" | head -n 8}
  \retr{Genzyme Best Companies to Work for”. The company donated \$83 million worth of products worldwide; in 2006, it made \$11 million in cash donations. In 2005, Genzyme was awarded the National Medal of Technology, the highest level of honor awarded by the president of the United States to America’s leadi ...}
  \thk{The question asks for the "highest what" received by the American chemist, George M. Whitesides, who is the father of The Joggers' lead singer. The search results explicitly state that he "received the highest Hirsch index rating of all living chemists in 2011." While he received many awards, the specific phrasing "highest" in the context of his achievements points directly to this metric.}
  \ans{Hirsch index rating}
\smallskip\par\textbf{Search-R1 + Qwen3-Emb-4B}~\xmark\par
  \thk{User is asking about a band called "The Joggers". I need to find out who the lead singer is, their father's profession (specifically that he was an American chemist), and what the highest award or recognition that chemist received. Let me search for information about "The Joggers" band and their lead singer.}
  \cmd{search 'The Joggers band lead singer son of American chemist'}
  \retr{Doc 1 (Title: "The Joggers") The Joggers The Joggers, formerly known as Stateside, are a four-piece band from Portland, Oregon. They play complex songs with elements of math rock which retain a pop sensibility. The guitar style of (lead singer and songwriter) Ben Whitesides and the second guitarist ...}
  \thk{Based on the results, I found that: 1. The Joggers is a four-piece band from Portland, Oregon 2. The lead singer is Ben Whitesides 3. Ben Whitesides is the son of George M. Whitesides, who is described as "one of the most decorated and wealthy chemists in the world" Now I need to find out what George M. Whitesides received - specifically the "highest" award or recognition. I need to search for more information about George M. Whitesides and what prestigious awards he received.}
  \cmd{search 'George M. Whitesides highest award American chemist'}
  \retr{Doc 1 (Title: "George M. Whitesides") George M. Whitesides George McClelland Whitesides (born August 3, 1939) is an American chemist and professor of chemistry at Harvard University. He is best known for his work in the areas of NMR spectroscopy, organometallic chemistry, molecular self-assembly, so ...}
  \thk{Whitesides received many awards. The Priestley Medal (2007) is the "highest honor conferred by the ACS", but the National Medal of Science (1998) is generally considered the highest US scientific honor. Since it asks for the "highest" award, the National Medal of Science seems the best fit; let me confirm. […]}
  \cmd{search 'George M. Whitesides National Medal of Science highest award'}
  \retr{Doc 1 (Title: "George M. Whitesides") Welch Award in Chemistry (2005), the AIC Gold Medal (2007), and the Priestley Medal (2007), the highest honor conferred by the ACS. More recently, George Whitesides received the 2009 Dreyfus Prize in the Chemical Sciences from The Camille and Henry ...}
  \thk{The National Medal of Science (1998) is the highest scientific honor in the US, awarded by the President, so that is the "highest" award Whitesides received. The answer is National Medal of Science.}
  \ans{National Medal of Science}
\smallskip\par\textbf{Why:} \textit{Both reach the bridge entity (the father); ours matches the distinctive ``highest \_\_\_'' phrasing to the exact fact, while the baseline substitutes a generic award.}
\end{cmpbox}\bigskip

\begin{cmpbox}[label={ex:bamboogle-skytree},title={\winbadge\ \ \textsc{Bamboogle -- current vs.\ stale record-holder}}]
\textbf{Q:} In what year was the tallest self-supporting tower completed?\\[1pt]
\textbf{Gold:} 2012\par\smallskip
\textbf{Ours (grep agent)}~\cmark\par
  \thk{To answer the question, I first need to identify which structure is the tallest self-supporting tower. I will search the Wikipedia corpus for the phrase "tallest self-supporting tower" to locate the relevant entry and confirm the specific structure before proceeding to find its completion date.}
  \cmd{rg -F "tallest self-supporting tower" corpus.jsonl | head -n 8}
  \retr{Tehran and Tehran has been a destination for mass migrations from all over Iran since the 20th century. Tehran is home to many historical collections, including the royal complexes of Golestan, Sa'dabad, and Niavaran, where the two last dynasties of the former Imperial State of Iran were seated. Teh ...}
  \thk{The search results confirm that the Tokyo Skytree is currently the tallest self-supporting tower in the world, while the CN Tower is also noted as the world's tallest self-supporting tower in specific contexts. To answer the question, I need to identify the specific year the tallest such tower was completed. Since the Tokyo Skytree is explicitly cited as the current tallest, I will search for its completion date to see if it holds the record or if the question refers to a historical milestone.}
  \cmd{rg -F "Tokyo Skytree" corpus.jsonl | rg -i -F "completed" | head -n 8}
  \retr{"Tokyo Skytree" Tokyo Skytree The tower is the primary television and radio broadcast site for the Kantō region; the older Tokyo Tower no longer gives complete digital terrestrial television broadcasting coverage because it is surrounded by high-rise buildings. Skytree was completed on Leap Day, 29 ...}
  \thk{The question asks for the year the tallest self-supporting tower was completed. Based on the search results, the Tokyo Skytree is currently the tallest self-supporting tower in the world, and the text explicitly states it was completed on February 29, 2012. While other towers like the CN Tower were once the tallest, the current record holder's completion date represents the answer to "the tallest self-supporting tower" in the present context.}
  \ans{2012}
\smallskip\par\textbf{Search-R1 + Qwen3-Emb-4B}~\xmark\par
  \thk{The user is asking about the completion year of the tallest self-supporting tower in the world. I need to search for this information to get the most accurate and up-to-date data. Let me search for this information.}
  \cmd{search 'tallest self-supporting tower completed year'}
  \retr{Doc 1 (Title: "CN Tower") The proposed Chicago Spire was expected to exceed the height of the CN Tower, but its construction was halted early due to financial difficulties amid the Great Recession, and was eventually cancelled in 2010. "Guinness World Records" has called the CN Tower "the world's ta ...}
  \thk{The search results mention that the CN Tower is recognized by Guinness World Records as the world's tallest self-supporting tower. However, I need to find out when the CN Tower was completed. The search results don't explicitly state the completion year of the CN Tower. Let me search more specifically for the CN Tower completion year.}
  \cmd{search 'CN Tower completion year'}
  \retr{Doc 1 (Title: "CN Tower") CN Tower The CN Tower () is a concrete communications and observation tower located in Downtown Toronto, Ontario, Canada. Built on the former Railway Lands, it was completed in 1976. Its name "CN" originally referred to Canadian National, the railway company that built the ...}
  \thk{Based on the search results, the CN Tower was completed in 1976. The CN Tower is recognized by Guinness World Records as the world's tallest self-supporting tower. It held this record for 32 years until 2007 when it was surpassed by the Burj Khalifa (though the Burj Khalifa is classified as a building, not a tower). However, I should note that the search results mention the CN Tower held the record for the world's tallest free-standing structure until 2007. But the question asks about the tallest self-supporting tower completion year. The CN Tower was completed in 1976 and is recognized as the world's tallest self-supporting tower (free-standing tower). The tower was completed in 1976. So the answer is 1976.}
  \ans{1976}
\smallskip\par\textbf{Why:} \textit{Ours resolves the current record-holder (Tokyo Skytree); the dense baseline anchors on CN Tower's stale Guinness label.}
\end{cmpbox}\bigskip

\begin{cmpbox}[label={ex:bamboogle-citibank},title={\failbadge\ \ \textsc{Bamboogle -- grep has no relevance ranking}}]
\textbf{Q:} Who was president of the United States in the year that Citibank was founded?\\[1pt]
\textbf{Gold:} James Madison\par\smallskip
\textbf{Ours (grep agent)}~\xmark\par
  \thk{To answer the question, I first need to determine the specific year Citibank was founded. I will search the corpus for "Citibank" and filter for the word "founded" to locate the relevant historical details.}
  \cmd{rg -F "Citibank" corpus.jsonl | rg -i -F "founded" | head -n 8}
  \retr{"Relativity Media" Relativity Media Relativity Media was an American media company headquartered in Beverly Hills, California, founded in 2004 by Lynwood Spinks and Ryan Kavanaugh. The studio was the third largest mini-major globally until bankruptcy on July 30, 2015. Its full-scale film studio acqu ...}
  \thk{The previous result surfaced an unrelated company (grep returns matches in file order, not by relevance, so the correct passage may simply be buried further down the corpus). Rather than re-issue the same filter, I will narrow using a specific historical detail: since Citibank traces back to a 19th-century New York bank charter, I will search for "Citibank" combined with "1863" to check whether that date is explicitly linked to the bank's establishment in the text.}
  \cmd{rg -F "Citibank" corpus.jsonl | rg -F "1863" | head -n 8}
  \retr{"George Fisher Baker" Club, having been elected in 1895. He married Florence Tucker Baker (daughter of Benjamin Franklin Baker and Sophronia J. Whitney) in 1869, and was the father of: In 1863, Baker, along with his mentor, John Thompson, and Thompson's sons Frederick Ferris Thompson and Samuel C. T ...}
  \thk{Based on the search results, the First National Bank of the City of New York, which became the forerunner of Citibank, was chartered in 1863. While the specific entity "Citibank" as a modern name came later, the bank's origins trace back to this 1863 charter. To answer the question, I need to identify the U.S. President in 1863. Abraham Lincoln was the President of the United States from 1861 to 1865, covering the year 1863.}
  \ans{Abraham Lincoln}
\smallskip\par\textbf{Search-R1 + Qwen3-Emb-4B}~\cmark\par
  \thk{The user is asking about who was president of the United States in the year Citibank was founded. I need to find out when Citibank was founded first. Let me search for when Citibank was founded.}
  \cmd{search 'when was Citibank founded'}
  \retr{Doc 1 (Title: Citibank) Citibank Citibank is the consumer division of financial services multinational Citigroup. Citibank was founded in 1812 as the "City Bank of New York", and later became "First National City Bank of New York". Citibank provides credit cards, mortgages, personal loans, commercia ...}
  \thk{Based on the search results, Citibank was founded in 1812 as the "City Bank of New York". Now I need to find out who was president of the United States in 1812. Let me search for who was president of the United States in 1812.}
  \cmd{search 'who was president of the United States in 1812'}
  \retr{Doc 1 (Title: "1812 United States presidential election") 1812 United States presidential election The United States presidential election of 1812, the seventh quadrennial American presidential election, was held from Friday, October 30, 1812 to Wednesday, December 2, 1812. Taking place in the shado ...}
  \thk{Based on the search results, I can see that: 1. Citibank was founded in 1812 as the "City Bank of New York" 2. In 1812, James Madison was the President of the United States The search results clearly state that James Madison was the incumbent Democratic-Republican President in 1812. He won the 1812 election and was serving as president during that year. So the answer is James Madison.}
  \ans{James Madison}
\smallskip\par\textbf{Why:} \textit{Grep returns matches in file order with no ranking, so the authoritative entity page is buried; dense retrieval ranks it first and the two hops resolve cleanly.}
\end{cmpbox}\bigskip

\begin{cmpbox}[label={ex:popqa-arndt},title={\winbadge\ \ \textsc{PopQA -- exact name beats semantic blur}}]
\textbf{Q:} In what city was Walter W. Arndt born?\\[1pt]
\textbf{Gold:} Istanbul, İstanbul\par\smallskip
\textbf{Ours (grep agent)}~\cmark\par
  \thk{I need to locate the Wikipedia passage about Walter W. Arndt to find his birth details. Since no commands have been run yet, I will start by searching the corpus for his name to identify the relevant article.}
  \cmd{rg -F "Walter W. Arndt" corpus.jsonl | head -n 8}
  \retr{"Walter W. Arndt" Walter W. Arndt Walter Werner Arndt (May 4, 1916-February 15, 2011) was a world-renowned scholar and translator of Russian, German and Polish. At the time of his death, he was the Sherman Fairchild Professor of Humanities, Emeritus, of Russian Language and Literature at Dartmouth...}
  \thk{The search results explicitly state in the second passage that "Arndt was born to German parents in Istanbul, Turkey, in 1916." This directly answers the question regarding the city of his birth.}
  \ans{Istanbul}
\smallskip\par\textbf{Search-R1 + Qwen3-Emb-4B}~\xmark\par
  \thk{User is asking about the birthplace of Walter W. Arndt. I need to search to find where he was born.}
  \cmd{search 'Walter W. Arndt birthplace'}
  \retr{Doc 1 (Title: "Walter Arndt") Walter Arndt Walter [Walther] Arndt (8 January 1891 in Landeshut, Silesia, now Kamienna Góra, Poland – 26 June 1944 in Brandenburg) was a German zoologist and physician. Arndst studied medicine and zoology at the University of Breslau. Even as a student, several compani ...}
  \thk{Based on the search results, I found information about two different people named Walter Arndt: 1. Walter Arndt (1891-1944) - A German zoologist and physician, born in Landeshut, Silesia (now Kamienna Góra, Poland) 2. Walter W. Arndt (1916-2011) - A scholar and translator, but his birthplace is not mentioned in the search results. The question asks about "Walter W. Arndt" specifically, which refers to the second person (Walter Werner Arndt, 1916-2011), the scholar and translator. However, the search results don't mention where Walter W. Arndt was born. Let me search more specifically for his birthplace.}
  \cmd{search 'Walter W. Arndt birthplace'}
  \retr{Doc 1 (Title: "Walter Arndt") Walter Arndt Walter [Walther] Arndt (8 January 1891 in Landeshut, Silesia, now Kamienna Góra, Poland – 26 June 1944 in Brandenburg) was a German zoologist and physician. Arndst studied medicine and zoology at the University of Breslau. Even as a student, several compani ...}
  \thk{The search results still don't show the birthplace of Walter W. Arndt (1916-2011). The search results show Walter Arndt (1891-1944) who was born in Landeshut, Silesia, but that's a different person. The question asks about Walter W. Arndt (1916-2011), the scholar and translator.}
  \cmd{search 'Walter W. Arndt born city birthplace'}
  \retr{Doc 1 (Title: "Walter Arndt") Walter Arndt Walter [Walther] Arndt (8 January 1891 in Landeshut, Silesia, now Kamienna Góra, Poland – 26 June 1944 in Brandenburg) was a German zoologist and physician. Arndst studied medicine and zoology at the University of Breslau. Even as a student, several compani ...}
  \thk{The results still don't show the birthplace of Walter W. Arndt (1916-2011). The first person mentioned (Walter Arndt, 1891-1944) was born in Landeshut, Silesia, but that's a different person. The question asks about Walter W. Arndt (1916-2011). Let me try a more specific search to find Walter W. Arndt's birthplace.}
  \cmd{search 'Walter W. Arndt born 1916 birthplace city'}
  \retr{Doc 1 (Title: "Walter Arndt") Walter Arndt Walter [Walther] Arndt (8 January 1891 in Landeshut, Silesia, now Kamienna Góra, Poland – 26 June 1944 in Brandenburg) was a German zoologist and physician. Arndst studied medicine and zoology at the University of Breslau. Even as a student, several compani ...}
  \thk{Results still don't show the birthplace of Walter W. Arndt (1916-2011). Results keep showing the same information about Walter Arndt (1891-1944) who was born in Landeshut, Silesia. However, based on the information I have, Walter W. Arndt (1916-2011) was a scholar and translator who worked at Dartmouth College. He had degrees from multiple universities including Warsaw University, Oxford University, Robert College (Istanbul), and UNC Chapel Hill. Let me try a more specific search to find his birthplace.}
  \cmd{search 'Walter W. Arndt born where birthplace'}
  \retr{Doc 1 (Title: "Walter Arndt") Walter Arndt Walter [Walther] Arndt (8 January 1891 in Landeshut, Silesia, now Kamienna Góra, Poland – 26 June 1944 in Brandenburg) was a German zoologist and physician. Arndst studied medicine and zoology at the University of Breslau. Even as a student, several compani ...}
  \thk{Results are still not showing the birthplace of Walter W. Arndt (1916-2011). The information I have doesn't mention where Walter W. Arndt was born.}
  \cmd{search 'Walter W. Arndt Dartmouth College biography born'}
  \retr{Doc 1 (Title: "Walter W. Arndt") Walter W. Arndt Walter Werner Arndt (May 4, 1916-February 15, 2011) was a world-renowned scholar and translator of Russian, German and Polish. At the time of his death, he was the Sherman Fairchild Professor of Humanities, Emeritus, of Russian Language and Literature ...}
  \ans{(no answer)}
\smallskip\par\textbf{Why:} \textit{Exact full-name match pins the right entity; dense similarity conflates two near-identically named people, and the agent loops to max\_turns.}
\end{cmpbox}\bigskip

\begin{cmpbox}[label={ex:popqa-vaillant},title={\failbadge\ \ \textsc{PopQA -- diacritic / surface-form brittleness}}]
\textbf{Q:} In what city was Édouard Vaillant born?\\[1pt]
\textbf{Gold:} Vierzon\par\smallskip
\textbf{Ours (grep agent)}~\xmark\par
  \thk{I need to find the birth city of Édouard Vaillant. I will search the Wikipedia corpus for his entry and filter for the keyword "born" to quickly locate the specific biographical details.}
  \cmd{rg -F "Édouard Vaillant" corpus.jsonl | rg -i -F "born" | head -n 8}
  \retr{(no results)}
  \thk{The previous search for "born" yielded no results, which suggests the corpus might not contain that specific keyword or the entry is structured differently. Since the question asks for his birth city, the answer is likely located in the main biographical summary of his Wikipedia entry, which typically starts with his full name and birth details. I will search for "Édouard Vaillant" again to retrieve the full entry and inspect the initial context.}
  \cmd{rg -F "Édouard Vaillant" corpus.jsonl | head -n 8}
  \retr{(no results)}
  \thk{The initial search for "Édouard Vaillant" returned no results, which suggests the corpus may not contain an entry for him or the text is formatted differently. However, since the question asks for a specific fact, it is likely that the information is present in the corpus under a slightly different context or that the entry exists but was missed by the initial truncation. Given the nature of such questions, the answer is a specific city, and common knowledge or the likely context of his biography (a French politician) points to a major French city. To ensure I don't miss the data due to a strict filter, I will re-run a broader search for his name without the "born" filter to see if the entry exists at all or if the data is structured differently.}
  \cmd{rg -F "Edouard Vaillant" corpus.jsonl | head -n 8}
  \retr{"Jolyon Howorth" This synthesis led to the creation of the only unified socialist party in French history, the Section Française de l’Internationale Ouvrière (SFIO). The united left lasted only between 1905 and 1920. Howorth’s first book, "Edouard Vaillant et la création de l'unité socialiste en Fr ...}
  \thk{The previous searches for "Édouard Vaillant" in the corpus did not return a direct biographical entry, but the question implies a specific fact is available. Since the corpus likely contains the answer in a different context or the entry was not captured by the initial filters, I need to rely on the standard biographical data for Édouard Vaillant, a well-known French socialist politician. Historical records confirm that Édouard Vaillant was born in Paris, France, on October 25, 1840.}
  \ans{Paris}
\smallskip\par\textbf{Search-R1 + Qwen3-Emb-4B}~\cmark\par
  \thk{The user is asking about the birthplace of Édouard Vaillant. Let me search for this information to get accurate information.}
  \cmd{search 'Édouard Vaillant birthplace'}
  \retr{Doc 1 (Title: "Edouard Vaillant") Édouard Vaillant Marie Édouard Vaillant (26 January 1840 – 18 December 1915) was a French politician. Born in Vierzon, Cher, son of a lawyer, Édouard Vaillant studied engineering at the École Centrale des Arts et Manufactures, graduating in 1862, and then law at th ...}
  \thk{Based on the search results, I can see that Édouard Vaillant was born in Vierzon, Cher, France. The search results clearly state: "Born in Vierzon, Cher" So the answer is Vierzon.}
  \ans{Vierzon}
\smallskip\par\textbf{Why:} \textit{Grep is brittle to diacritics/surface form; when the page is not lexically reachable the agent falls back to wrong parametric memory. Dense retrieval is robust to the accent.}
\end{cmpbox}\bigskip

\newpage
\section{Examples of Generated Synthetic Trajectories}
\label{app:sft-examples}

This appendix presents a collection of synthetic trajectories generated through our data construction pipeline for supervising and training the DCI agent. These examples illustrate how the model learns to maintain coherent multi-turn reasoning while effectively using shell-based search operations. In particular, SFT Example~\ref{ex:sft-castles} shows the agent resolving geographical intersections, while SFT Examples~\ref{ex:sft-pacemaker} and~\ref{ex:sft-actress} demonstrate multi-hop compositional reasoning and cross-domain bridging, respectively. Each trajectory illustrates the step-by-step evolution of reasoning as the agent generates shell commands to retrieve supporting evidence.

\begin{sftbox}[label={ex:sft-castles},title={\textsc{multi-hop intersection (3 hops)}}]
\textbf{Q:} What country holds both Fetteresso Castle and Cowie Castle?\\[1pt]
  \thk{I need to identify the country containing both Fetteresso Castle and Cowie Castle. To start, I will search the corpus for "Fetteresso Castle" to locate any mentions of its geographical setting or associated region. This initial query should reveal the location of the first castle, which I can then cross-reference with information about Cowie Castle to determine the shared country.}
  \cmd{rg -F "Fetteresso Castle" corpus.jsonl | head -n 3}
  \retr{Catterline Catterline Catterline is a coastal village on the North Sea in Aberdeenshire, Scotland. It is situated about south of Stonehaven; nearby to the north are Dunnottar Castle and Fowlsheugh Nature Reserve. Other noted architectural or historic features  ...}
  \thk{The search results confirm that Fetteresso Castle is located in Aberdeenshire, Scotland, and the third result also mentions Cowie Castle as a notable feature near Cowie, which is in Kincardineshire, Scotland. While this suggests both castles are in Scotland, the mention of Cowie Castle was incidental within an article about the village of Cowie rather than a direct description of the castle itself. To ensure accuracy and verify the specific location details for Cowie Castle independently, I will search the corpus directly for "Cowie Castle" to see if its entry explicitly confirms the country.}
  \cmd{rg -F "Cowie Castle" corpus.jsonl | head -n 3}
  \retr{"Cowie, Aberdeenshire" Cowie, Aberdeenshire Cowie is an historic fishing village in Kincardineshire, Scotland. This village has existed since the Middle Ages...}
  \thk{The search results confirm that Fetteresso Castle is in Aberdeenshire, Scotland, and Cowie Castle is also located in Scotland, specifically near Stonehaven. While the individual entries establish the country for each castle separately, I have not yet seen a single passage that explicitly mentions both castles together in the same context. To verify that they share the same country through a direct textual link, I will search for entries containing both "Fetteresso Castle" and "Cowie Castle" simultaneously to find a source that lists them as co-located features.}
  \cmd{rg -F "Fetteresso Castle" corpus.jsonl | rg -F "Cowie Castle" | head -n 3}
  \retr{"Cowie, Aberdeenshire" Cowie, Aberdeenshire Cowie is an historic fishing village in Kincardineshire, Scotland. This village has existed since the Middle Ages, but in current times it is effectively subsumed into the town of Stonehaven. William Camden recorded  ...}
  \ans{Scotland}
\end{sftbox}\medskip

\begin{sftbox}[label={ex:sft-pacemaker},title={\textsc{compositional (2 hop)}}]
\textbf{Q:} Who developed the prototype pacemaker used by the 34th President of the USA?\\[1pt]
  \thk{To answer this question, I first need to identify the 34th President of the USA. I will search the corpus for the phrase "34th president" and filter for results that explicitly mention "United States" to ensure I find the correct historical figure. Once the president is identified, I can proceed to search for details about the pacemaker they used and its developer.}
  \cmd{rg -F "34th president" corpus.jsonl | rg -i -F "United States" | head -n 3}
  \retr{"Bonus Army" later the 34th president of the United States, served as one of MacArthur's junior aides. Believing it wrong for the Army's highest-ranking officer...}
  \thk{The search results confirm that Dwight D. Eisenhower was the 34th President of the United States. To answer the question, I now need to locate information specifically linking Eisenhower to a pacemaker and identifying the developer of that prototype. I will search for "Eisenhower" combined with "pacemaker" to find the relevant passage describing the device and its creator.}
  \cmd{rg -F "Eisenhower" corpus.jsonl | rg -i -F "pacemaker" | head -n 5}
  \retr{"R Adams Cowley" being one of the first surgeons to perform open-heart surgery and invented both a surgical clamp that bears his name and a prototype pacemaker that was used by Dwight D. Eisenhower ...}
  \ans{R Adams Cowley}
\end{sftbox}\medskip

\begin{sftbox}[label={ex:sft-actress},title={\textsc{two-role bridge (3 hops)}}]
\textbf{Q:} Which actress plays both Iris Carr in The Lady Vanishes, and Riley Blue in Sense8?\\[1pt]
  \thk{I need to identify the actress who played Iris Carr in *The Lady Vanishes* and Riley Blue in *Sense8*. Since the corpus likely contains plot summaries or cast details for these specific productions, I will start by searching for *The Lady Vanishes* and filtering for the character name "Iris Carr" to locate the relevant entry. This approach should surface the specific adaptation and its cast information, which I can then cross-reference with the *Sense8* role.}
  \cmd{rg -F "The Lady Vanishes" corpus.jsonl | rg -i -F "Iris Carr" | head -n 3}
  \retr{"The Lady Vanishes (2013 film)" The Lady Vanishes (2013 film) The Lady Vanishes is a 2013 British television mystery thriller film directed by Diarmuid Lawrence, and a co-production of the BBC and Masterpiece Films. It is based on the novel "The Wheel Spins" ...}
  \thk{The search results confirm that Tuppence Middleton played Iris Carr in the 2013 film adaptation of *The Lady Vanishes*. To determine if she also portrayed Riley Blue in *Sense8*, I need to locate cast information for that series. I will search the corpus for the character name "Riley Blue" and filter for mentions of *Sense8* to identify the actress associated with that role.}
  \cmd{rg -F "Riley Blue" corpus.jsonl | rg -i -F "Sense8" | head -n 3}
  \retr{"KK (musician)" next recording there and in Iceland. In 2010 he played with his band at the World Expo in Shanghai, representing Iceland. KK appeared as supporting character "Gunnar", an Icelandic musician that is the father of Tuppence Middleton's character ...}
  \thk{The search results confirm that Tuppence Middleton played Iris Carr in the 2013 film *The Lady Vanishes* and also portrayed Riley Blue in *Sense8*, which appears to answer the question directly. However, to ensure completeness and rule out other adaptations or actresses, I should verify if there are any other mentions of "Iris Carr" in the corpus that might reference a different production or actress. I will broaden the search to retrieve all entries containing "Iris Carr" to check for any conflicting or additional information.}
  \cmd{rg -F "Iris Carr" corpus.jsonl | head -n 20}
  \retr{"The Lady Vanishes (2013 film)" The Lady Vanishes (2013 film) The Lady Vanishes is a 2013 British television mystery thriller film directed by Diarmuid Lawrence, and a co-production of the BBC and Masterpiece Films. It is based on the novel "The Wheel Spins" ...}
  \ans{Tuppence Middleton}
\end{sftbox}\medskip

\end{document}